%% file: main.tex
\documentclass[11pt]{article}

\usepackage[preprint]{acl}

\usepackage{times}
\usepackage{latexsym}
\usepackage[T1]{fontenc}
\usepackage[utf8]{inputenc}
\usepackage{microtype}
\usepackage{inconsolata}
\usepackage{graphicx}
\usepackage{xurl}  %

\usepackage{booktabs}
\usepackage{multirow}
\usepackage{amsmath}
\usepackage{amssymb}
\usepackage{xspace}
\usepackage{xcolor}
\usepackage{tikz}
\usetikzlibrary{positioning,arrows.meta,fit,backgrounds,calc,shapes.misc}
\usepackage{pgfplots}
\pgfplotsset{compat=1.17}
\usepackage{tcolorbox}
\tcbuselibrary{skins}

\definecolor{promptframe}{HTML}{2E4A62}
\definecolor{promptback}{HTML}{F7F9FB}
\definecolor{promptvar}{HTML}{8A3324}
\definecolor{exampleframe}{HTML}{23645E}
\definecolor{exampleback}{HTML}{F6FAF9}

\newtcolorbox{promptbox}[1]{%
  enhanced,
  colback=promptback,
  colframe=promptframe,
  colbacktitle=promptframe,
  coltitle=white,
  fonttitle=\bfseries\small\hypersetup{linkcolor=white},
  fontupper=\footnotesize,
  boxrule=0.6pt,
  arc=2.5pt,
  toptitle=2.5pt,
  bottomtitle=2.5pt,
  left=7pt,
  right=7pt,
  top=5pt,
  bottom=5pt,
  before upper={\setlength{\parskip}{2.5pt}},
  title={#1}
}

\newtcolorbox{examplebox}[1]{%
  enhanced,
  colback=exampleback,
  colframe=exampleframe,
  colbacktitle=exampleframe,
  coltitle=white,
  fonttitle=\bfseries\small\hypersetup{linkcolor=white},
  fontupper=\footnotesize,
  boxrule=0.6pt,
  arc=2.5pt,
  toptitle=2.5pt,
  bottomtitle=2.5pt,
  left=7pt,
  right=7pt,
  top=5pt,
  bottom=5pt,
  before upper={\setlength{\parskip}{2.5pt}},
  title={#1}
}

\newcommand{\prole}[1]{\textbf{\textcolor{promptframe}{#1}}}
\newcommand{\erole}[1]{\textbf{\textcolor{exampleframe}{#1}}}
\newcommand{\pvar}[1]{\texttt{\textcolor{promptvar}{\{#1\}}}}
\newcommand{\tcbsep}{\par\vspace{3pt}\noindent\textcolor{black!18}{\rule{\linewidth}{0.4pt}}\par\vspace{3pt}}

\newif\ifdraft
\draftfalse
\ifdraft
  
  \newcommand{\pend}[1]{\textcolor{red}{[#1]}}
  
  \newcommand{\draftnote}[1]{\textcolor{red}{\textbf{[Draft note:} #1\textbf{]}}}
\else
  
  \newcommand{\pend}[1]{}
  
  \newcommand{\draftnote}[1]{}
\fi

\newcommand{\data}{Any-DP\xspace}
\newcommand{\sone}{System-1\xspace}
\newcommand{\stwo}{System-2\xspace}

\newcommand{\ours}{AnyS1\xspace}
\newcommand{\gmain}{AnyS1\textsubscript{warm}\xspace}
\newcommand{\gzero}{AnyS1\textsubscript{cold}\xspace}
\newcommand{\syn}{S1-syn\xspace}

\newcommand{\none}{\textit{none of the above}\xspace}

\title{Thinking Fast, Anywhere: One System-1 for the Decision Points\\of Any Agent Workflow}

\author{Rui Li$^{1}$ \quad Liyang He$^{1}$ \quad Zheng Zhang$^{2}$ \quad Zhenya Huang$^{1}$ \quad Qi Liu$^{1}$\\
  $^{1}$University of Science and Technology of China \quad $^{2}$Nanyang Technological University\\
  \texttt{ruili2000@mail.ustc.edu.cn}\\
  \url{https://middle-chunjie.github.io/AnyS1/}}

\hypersetup{pdftitle={Thinking Fast, Anywhere: One System-1 for the Decision Points of Any Agent Workflow},pdfauthor={Rui Li, Liyang He, Zheng Zhang, Zhenya Huang, Qi Liu}}
\begin{document}
\maketitle

\input{sections/abstract}

\input{sections/introduction}
\input{sections/related}
\input{sections/method}
\input{sections/experiments}
\input{sections/conclusion}

\input{sections/limitations}

\bibliography{references}

\appendix
\input{sections/appendix}

\end{document}

%% file: sections/abstract.tex
\begin{abstract}
Autonomous LLM agents incur substantial cost and latency by invoking a large model (\stwo) at every step, even when many intermediate actions choose among a small set of known candidates.
Existing dual-system agents offload such decisions to a lightweight \sone model, but rely on workflow-specific hooks authored per environment.
We present \ours, a framework that turns human-authored procedural skills—whose instructions naturally specify where choices arise and which options apply—into a shared, domain-general source of decision points.
Offline, \ours mines grounded enumerable decisions across over 1,000 public skills, gathers execution-grounded trajectory states inside isolated containers to construct \data, and trains a single calibrated 4B pointer \sone with label smoothing.
At run time, compiled workflows resolve routine decisions in one forward pass and fall back to \stwo when confidence falls below a calibrated threshold.
Across three unseen benchmarks and distinct \stwo model families, \ours preserves task reliability while cutting episode cost by 75--91\% and model-call time by 63--92\% relative to an uncompiled \stwo agent, and maintains performance under zero-shot threshold transfer.
\end{abstract}

%% file: sections/introduction.tex
\section{Introduction}
\label{sec:intro}

Driven by advances in reasoning and tool use, large language models (LLMs) have evolved into autonomous agents capable of complex, long-horizon tasks \citep{kwa2025longtasks}—resolving repository-level software issues \citep{yang2024sweagent}, operating desktop and web interfaces \citep{xie2024osworld,zhou2024webarena}, and handling policy-constrained customer workflows \citep{yao2024taubench,barres2025tau2}.
Yet mainstream agent architectures invoke a large model (\stwo) at every step over a growing trajectory \citep{yao2023react,wang2024openhands}, causing inference latency and cost to escalate rapidly \citep{kapoor2024agentsmatter}—even though many intermediate steps choose among a small set of known options rather than requiring open-ended generation (\S\ref{sec:rq3}).

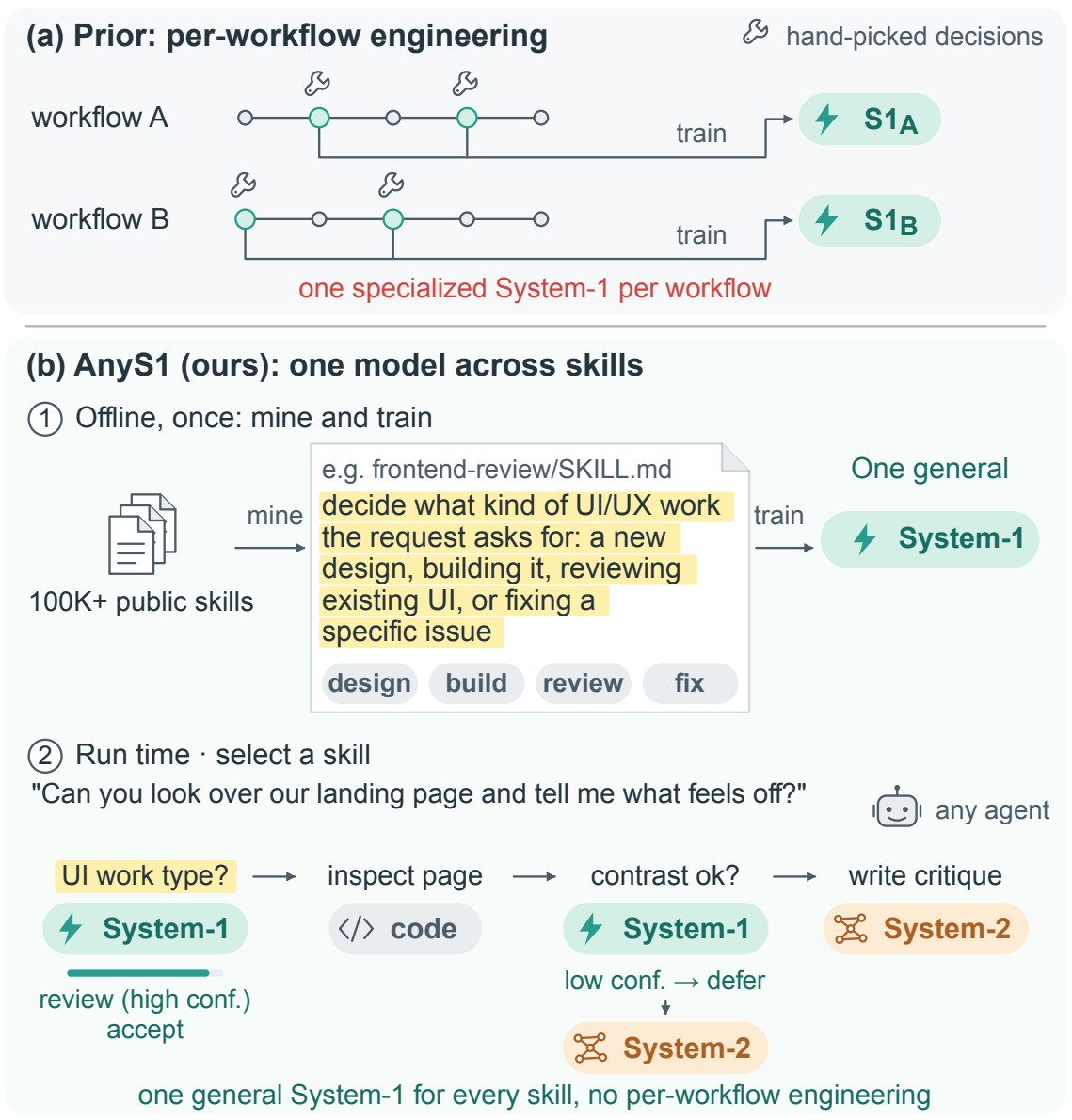
\begin{figure}[t]
  \centering
  \input{figures/fig_motivation}
  \caption{Prior dual-system agents versus \ours. (a)~Each prior agent adapts \sone to one fixed workflow: engineers hand-pick its decision points and train a \sone specialized to them. (b)~\ours mines decision points from public skills once, offline, and trains one general \sone; any agent that selects a skill then executes it as a dual-system workflow.}
  \label{fig:motivation}
\end{figure}

Typed decision models offer a fast alternative for such bounded choices \citep{typesafe2026jev,kev2026}: given the current state, a question, and candidate options, they output an option distribution in a single forward pass, acting as a fast, intuitive \sone alongside \stwo for planning, open generation, and uncertain fallbacks \citep{kahneman2011thinking}.
While recent dual-system agents use this collaboration to substantially cut decision latency and cost in search and specialized tool harnesses \citep{cao2026searchjev,zhu2026jevagentharness,wu2026reflex}, they remain restricted to decision points manually pre-defined for a single fixed workflow, collecting training supervision exclusively at those hard-coded hooks (Figure~\ref{fig:motivation}(a)).
When a general agent faces diverse tasks governed by new workflows without pre-declared hooks, every judgment falls back to \stwo.

We observe that general agents increasingly acquire workflows through \emph{agent skills} (\texttt{SKILL.md}; \citealp{anthropic2025skills}), a paradigm uniting \emph{flexibility} with \emph{procedurality}.
Across public repositories of over a hundred thousand skills \citep{skillmd138k,wang2026skillcorpus}, an agent can dynamically load relevant skills on demand, while each skill specifies a structured procedure whose branches and checks implicitly define where to judge and among which options (Figure~\ref{fig:motivation}(b)).
Procedural skills thus offer a natural, workflow-agnostic source from which decision points can be mined automatically for dual-system execution.

Building on this insight, we propose \ours, a framework that mines actionable decision points from procedural skills and extends single-LLM execution into calibrated dual-system collaboration (Figure~\ref{fig:framework}).
Offline, \ours extracts verbatim-grounded enumerable decision points across over 1,000 public skills and executes synthesized tasks in isolated containers to build \data—17,865 typed decision points and 25,986 teacher-consensus execution states—training one 4B \sone pointer model with label smoothing for cross-workflow calibration.
At run time, compiled workflows run deterministic steps as code, keep open generation on \stwo, and score enumerable decisions with \sone in one forward pass: a confidence gate adopts millisecond \sone choices above a threshold (calibrated on unlabeled development runs in \gmain or transferred zero-shot in \gzero) and defers uncertain cases to \stwo.

We evaluate \ours end-to-end on three unseen agent benchmarks—SOPBench, ALFWorld, and WebShop—across two distinct \stwo backbones (DeepSeek-V4.1-Flash and GPT-6 Sol).
Relative to an uncompiled \stwo agent that reads skills directly, \gmain resolves most enumerable decisions in a single millisecond-scale pass, reducing episode cost by 75--91\% and cutting critical-path model time by 63--92\% across both \stwo backbones without sacrificing task success.
With zero target-workflow runs, directly transferring a single global threshold (\gzero) likewise preserves task reliability across all three unseen benchmarks, confirming that one skill-trained \sone generalizes across domains and \stwo families without retraining.
Controlled analyses further show that training on authentic container-executed states outperforms LLM-fabricated states by +18.8 points on held-out observations, while calibration and confidence gating prevent cascading search failures and capture nearly the full savings ceiling set by each workflow's enumerable-step share.

Our main contributions are summarized as follows:
(1)~\textbf{Skill-mined paradigm and dataset:} we generalize dual-system agents from manually hard-coded workflow hooks to decision points mined automatically from procedural skills, releasing \data across 1,072 public skills;
(2)~\textbf{Calibrated dual-system framework:} we propose \ours, pairing one label-smoothed 4B \sone pointer model with confidence deferral under zero-shot transfer (\gzero) or unlabeled development calibration (\gmain);
(3)~\textbf{End-to-end validation and analysis:} across three unseen benchmarks and two \stwo families, we demonstrate substantial cost and latency reductions at preserved success, establish why authentic execution states and calibration are essential, and characterize the enumerable-cost ceiling.

%% file: figures/fig_motivation.tex
\includegraphics[width=\columnwidth]{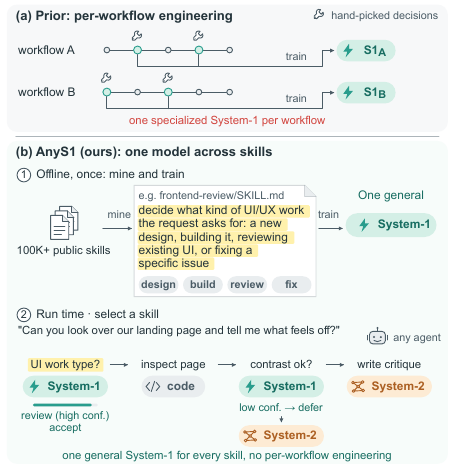}

%% file: sections/related.tex
\section{Related Work}
\label{sec:related}

\paragraph{Agent skills and workflow compilation.}
Agent skills package procedural knowledge as instructions and resources loaded on demand without weight updates \citep{anthropic2025skills}, extending executable skill libraries learned in open-ended play \citep{wang2023voyager}.
One line of work studies how agents retrieve, follow and internalize skills \citep{xia2026skillrl,lu2026skill0,li2026skillsbench,dong2026harmful,cho2026skillfollowing}, or synthesizes interaction trajectories and verifiable tasks from skill corpora to fine-tune generative policies \citep{zeng2024agenttuning,song2024eto,tan2026skt,wang2026skillgym,wang2026skillcorpus}; a complementary line compiles natural-language procedures into explicit programs \citep{khattab2024dspy}, turning SOPs into executable pseudo-code \citep{yu2026compile}, skills into runtime interfaces or state machines \citep{chen2026skvm,xu2026skillsmith,li2026hexis}, and recurring paths into scheduled workflows, deterministic rules or neurosymbolic programs \citep{xiang2026las,weber2026habits,liu2026harecap,suh2026nscu,mazza2026scope,wang2026abacus}.
However, prior skill-learning methods treat each skill as a unified prompt or trajectory-level target for a single generative model, while compiled workflows still rely on LLM calls, static rules or uncalibrated heuristic guards at semantic branches.
In contrast, \ours operates at the granularity of individual decisions inside skills: we mine enumerable decision points across diverse skills and harvest real execution states (\data) to train one calibrated \sone, turning compiled skill execution into a collaborative \sone--\stwo workflow.

\paragraph{Typed decision models and calibrated deferral.}
Inspired by dual-process theory \citep{kahneman2011thinking} and LLM cascades \citep{chen2023frugalgpt,ong2025routellm,xu2026gengate}, early fast--slow agents pair a small text generator or modular executor with a slow planner \citep{lin2023swiftsage,yin2024lumos,prasad2024adapt,saha2024system1x,su2024dualformer}, whereas typed decision models cast \sone as a non-generative scorer returning option probabilities in one forward pass.
Following commercial Jev \citep{typesafe2026jev}, open-weight models either attach pointer or decision heads to open LMs—for general choices (Kev; \citealp{kev2026}) or domain-specific agent states (SearchJev, \citealp{cao2026searchjev}; AgentDecision, \citealp{zhu2026jevagentharness})—or read debiased probabilities via cyclic permutations (AnyJev; \citealp{zhang2026anyjev}) to mitigate order sensitivity \citep{zheng2024mcq}.
Dual-system agents built on these models accelerate search \citep{cao2026searchjev} and tool use \citep{wu2026reflex,zhu2026jevagentharness} via selective deferral to an LLM \citep{geifman2017selective,sowietzki2026aria,hemadri2026r2v}, with recent studies analyzing delegation boundaries and biases \citep{vats2026goodjudgment,li2026readingright,lukassen2026cheap,sun2026typesafe}.
We generalize this paradigm beyond per-workflow manual hooks by mining decision points from skills and serving them with one label-smoothed \sone \citep{guo2017calibration,muller2019labelsmoothing} under unlabeled development calibration (\gmain) or zero-shot transfer (\gzero).

%% file: sections/method.tex
\section{\ours: One System-1 from Mined Decision Points}
\label{sec:method}

\begin{figure*}[t]
  \centering
  \input{figures/fig_framework}
  \caption{Overview of \ours on one decision point of a front-end skill. Offline (left), \data mines the decision point from \texttt{SKILL.md}, collects authentic states by executing synthesized tasks in containers, and keeps labels on which two order-reversed teacher passes agree, to train one general \sone (\S\ref{sec:mining}--\S\ref{sec:training}). Online (right), D steps run as code, G steps go to \stwo, and \sone answers E steps unless its confidence falls below $\tau$, in which case it defers to \stwo (\S\ref{sec:gate}). State text and probabilities are illustrative.}
  \label{fig:framework}
\end{figure*}
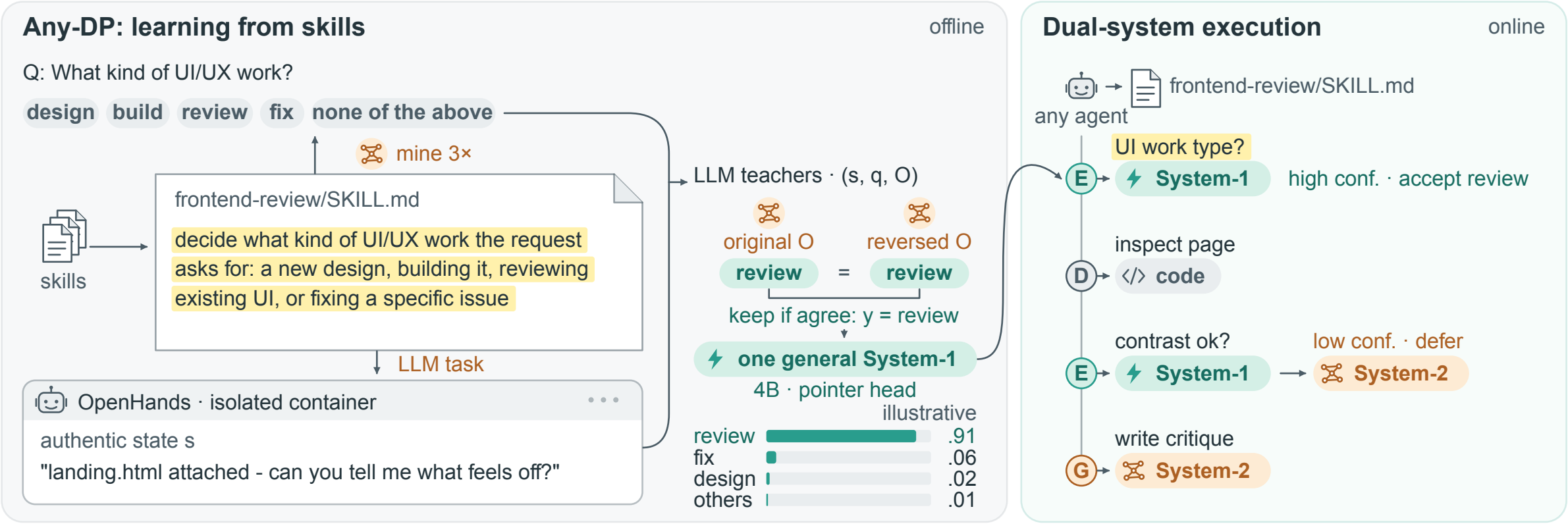

Figure~\ref{fig:framework} overviews \ours: after formalizing dual-system execution (\S\ref{sec:setup}), we mine decision points from skills (\S\ref{sec:mining}), collect consensus-labeled execution states (\S\ref{sec:states}) to train a calibrated 4B \sone (\S\ref{sec:training}), and gate runtime deferral (\S\ref{sec:gate}).
All offline LLM steps use DeepSeek-V4.1-Flash \citep{deepseek2026v4} (prompts in Appendix~\ref{app:data}).

\subsection{Problem Formulation}
\label{sec:setup}

\paragraph{From single- to dual-system execution.}
Consider an agent executing a task by following a skill over a multi-step \emph{episode}: at each step $t$, the agent observes the current state $s_t$ (accumulated context, tool outputs and user messages) and produces an action or decision.
A conventional single-system agent queries the LLM (\stwo) at every step to reason and generate the output.
To turn this into dual-system execution, we draw on dual-process theory \citep{kahneman2011thinking} and classify workflow steps by task type and decision complexity into three node types: a \emph{D node} (deterministic) is fully determined by structured state (e.g., a fixed tool call) and runs directly as code; a \emph{G node} (generative) produces open-ended output and stays with \stwo; and an \emph{E node} (enumerable) reads unstructured content in state $s$ and selects one option from a finite candidate set fixed in the skill or listable from $s$, making it suited to \sone delegation.
We formalize each skill-specified E node as a \emph{decision point} template $(q,O,d)$: a question $q$, options $O=\{o_1,\dots,o_n\}$ with one-sentence definitions $d(o)$, and \none; triggered at state $s$, it forms a runtime instance $x=(s,q,O,d)$.

\paragraph{System-1 and delegation.}
To resolve enumerable E nodes without the latency and cost of autoregressive generation, we introduce a lightweight typed decision model $f$ (\sone) that maps $x$ in one forward pass to a distribution $p_f(\cdot\mid x)$ over $O$, predicting $\hat y=\arg\max_{o\in O} p_f(o\mid x)$.
Because a compact $f$ has limited capacity and encounters unseen skills or ambiguous states where a wrong branch can derail downstream trajectories, we selectively delegate only high-confidence judgments to \sone: at each E node, a delegation rule $g(x)\in\{0,1\}$ decides whether to adopt $\hat y$ ($g(x)=1$) or fall back to \stwo ($g(x)=0$), where \emph{All-S2} ($g\equiv 0$) and \emph{All-S1} ($g\equiv 1$) correspond to relying solely on \stwo or \sone as the two extremes.

\paragraph{Objective.}
To formalize how selective delegation reduces inference cost while preserving task success, let $C$ and $C^{\mathrm{S2}}$ denote the per-episode costs of a dual-system configuration and All-S2.
Splitting $C$ into \stwo cost at G and E nodes and \sone cost gives
\begin{equation}
  \frac{C}{C^{\mathrm{S2}}} \;=\; \underbrace{\frac{C_G}{C^{\mathrm{S2}}}}_{\approx\,1-s_E} \;+\; \underbrace{\frac{C_E}{C^{\mathrm{S2}}}}_{\le\, s_E} \;+\; \underbrace{\frac{C_{\mathrm{S1}}}{C^{\mathrm{S2}}}}_{\kappa},
  \label{eq:cost}
\end{equation}
where $s_E$ is the share of All-S2 cost spent at E nodes and $\kappa$ is the \sone overhead (bounding savings by $s_E-\kappa$ when trajectories are unchanged).
We therefore learn a decision model $f$ that generalizes to unseen skills using supervision mined solely from skill corpora (\S\ref{sec:mining}--\S\ref{sec:training}), paired with a lightweight calibrated gate $g$ (\S\ref{sec:gate}), to remove as much E-node cost $C_E$ as possible while keeping success non-inferior to All-S2.

\subsection{Skill-Grounded Data Synthesis: \data}
\label{sec:mining}\label{sec:states}\label{sec:data}

To train a general \sone, we construct \data by first mining decision point templates $(q,O,d)$ from unstructured skill prose (\texttt{SKILL.md}) and then executing synthesized tasks in isolated containers to collect authentic runtime states $s$ and consensus labels $y\in O$.

\paragraph{Extracting decision points from skills.}
To ensure that mined decision points reflect genuine workflow branches, for each skill an LLM analyzer independently reads \texttt{SKILL.md} and its bundled documents three times, returning typed candidate decision points with options and a verbatim supporting quote verified via exact string matching against the skill text.
We then call an LLM to merge equivalent points across the three runs, record support $\sigma(p)\in\{1,2,3\}$, take the majority value per field, and write a one-sentence option definition $d(o)$ from the skill text to disambiguate identically named options.
We retain E nodes with $\sigma(p)\ge 2$ that are answerable from state alone, yielding grounded templates $(q,O,d)$.

\paragraph{Synthesizing executable tasks from skills.}
Extracted templates $(q,O,d)$ specify what to judge, yet training \sone also requires states $s$ that reflect genuine runtime contexts: asking an LLM to fabricate states directly yields clean prose summaries that fail to generalize to fragmented tool outputs along real trajectories.
To automatically elicit authentic deep-branch states across diverse skills, for each skill (split beforehand at the repository level into training and held-out sets to prevent leakage) an LLM synthesizes self-contained multi-stage tasks complete with input files, a reference solution, and a hidden verifier, paired with two cross-category distractor skills that require the agent to both select the right skill and advance across dependent stages; a task is retained only if, inside a network-isolated container, the verifier rejects the initial files and accepts the reference solution.

\paragraph{Trajectory execution and state labeling.}
To capture authentic interaction contexts, we use an OpenHands agent \citep{wang2024openhands} to execute each verified task in the isolated container, after which an LLM locator aligns the trajectory to the skill's decision points and extracts verbatim the preceding observations (task instructions, user messages, and raw tool outputs) at the step where each decision arises.
Because the executing agent's own actions may be suboptimal and single-pass LLM teachers exhibit option-order bias, we query two independent teacher passes per extracted state $s$ (reversing option order in the second pass) and retain only order-invariant consensus examples ($y^{1}=y^{2}$, setting $y=y^{1}$), providing clean supervision for calibrating \sone.

\begin{table}[t]
  \centering
  \input{tables/tab_data}
  \caption{Core statistics of \data across 1,072 public skills from SkillMD-138K (curation criteria, prompt templates, and worked example in Appendix~\ref{app:data}).}
  \label{tab:data}
\end{table}

\paragraph{Synthesized dataset \data.}
Running this pipeline on 1,072 public skills from SkillMD-138K \citep{skillmd138k}—spanning 16 categories such as web front-end, software engineering, and data analysis—yields \data, comprising 17,865 typed decision points, 2,846 container-verified tasks, and 25,986 consensus-labeled examples (Table~\ref{tab:data}).
Its distribution highlights two key properties: first, enumerable E nodes account for 64.5\% of all workflow nodes and 100\% of skills contain at least one E node, showing that bounded decisions are pervasive across procedural skills; second, 64.6\% of E nodes require contextual semantic inference rather than literal lookup (17.8\%), underscoring why static rules cannot replace a learned \sone.
Moreover, dual-teacher consensus reaches 89.2\%, with strict repository-level exclusion of all evaluation-related skills (Appendix~\ref{app:data}).

\subsection{Calibrated System-1 Training and Runtime Deferral}
\label{sec:training}\label{sec:gate}

\paragraph{Model architecture and calibrated objective.}
We build \sone by attaching an option-interaction pointer head to Qwen3.5-4B-Base \citep{qwen2026qwen35} and training it on \data.
Given a runtime instance $x=(s,q,O,d)$ with $O=\{o_1,\dots,o_n\}$, we serialize $x$ using structural delimiters into a single input sequence:
\begin{equation}
  \begin{aligned}
    \tilde{x} = \big[&\langle\mathrm{s}\rangle, s;\, \langle\mathrm{q}\rangle, q;\\[-2pt]
    &\big(\langle\mathrm{opt}\rangle, o_i{:}\,d(o_i), \langle/\mathrm{opt}\rangle\big)_{i=1}^{n};\, \langle\mathrm{dec}\rangle\big].
  \end{aligned}
  \label{eq:input}
\end{equation}
From one forward pass over $\tilde{x}$, this head passes the $\langle\mathrm{dec}\rangle$ hidden state and the $n$ $\langle/\mathrm{opt}\rangle$ hidden states through a position-free bidirectional self-attention layer so candidate options contextualize one another, then matches each option against $\langle\mathrm{dec}\rangle$ to output $p_f(o\mid x)$ via softmax (Appendix~\ref{app:training}).
Because runtime deferral relies on $p_f(o\mid x)$ to reflect true correctness likelihood on unseen skills, whereas hard-label cross-entropy drives overconfident predictions \citep{guo2017calibration}, we build calibration directly into training via label smoothing \citep{muller2019labelsmoothing}:
\begin{equation}
  \mathcal{L} = \sum_{(x,y)\in\mathcal{D}} \mathrm{CE}\big(p_f(\cdot\mid x),\,(1-\epsilon)\,e_y+\epsilon\,u_O\big),
  \label{eq:loss}
\end{equation}
where $e_y$ is the one-hot label, $u_O$ is uniform over $O$, and $\epsilon=0.1$.

\paragraph{Confidence-gated deferral.}
To prevent \sone errors on unfamiliar states from derailing downstream execution, the runtime gate delegates an E decision to \sone only when its maximum probability $c(x)=\max_{o\in O} p_f(o\mid x)$ clears a threshold $\tau$:
\begin{equation}
  g(x)=\mathbb{I}\big[c(x)\ge\tau\big],
  \label{eq:gate}
\end{equation}
where $\tau$ is the smallest threshold achieving at least 95\% agreement on calibration decisions with $c(x)\ge\tau$.
To support both brand-new workflows and workflows already in deployment, we determine $\tau$ under two regimes (fixed before any test run; Appendix~\ref{app:gate}):
(1)~\emph{Cold-start transfer (\gzero)}: to operate with zero target-task executions, \gzero fits a single global $\tau$ on 153 examples from held-out development skills and applies it unchanged to every benchmark;
(2)~\emph{Unlabeled development calibration (\gmain)}: to adapt $\tau$ to an active workflow without human annotation, \gmain executes 50 disjoint development episodes with \stwo while recording \sone's scores alongside \citep{zhu2026jevagentharness}, fitting $\tau$ against \stwo's decisions.

%% file: figures/fig_framework.tex
\includegraphics[width=\textwidth]{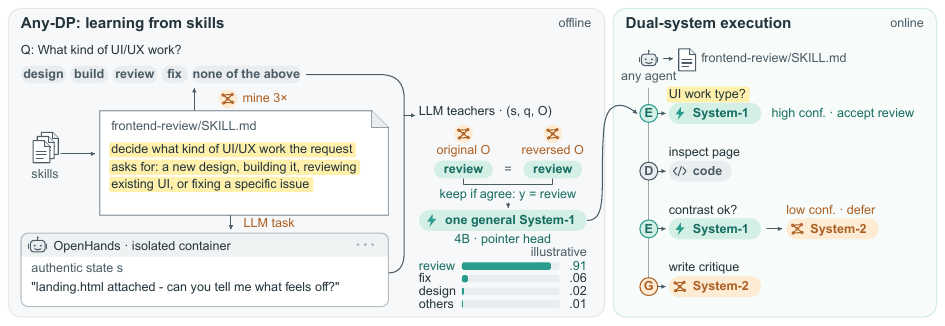}

%% file: tables/tab_data.tex
\footnotesize
\setlength{\tabcolsep}{4pt}
\begin{tabular}{@{}lr@{}}
\toprule
Public skills & 1,072 \\
Skill categories & 16 \\
Decision points (D / E / G, \%) & 17,865 (21.3 / 64.5 / 14.2) \\
Skills with $\ge$1 E node & 100\% \\
Synthesized tasks & 2,846 \\
Consensus states $|\mathcal{D}|$ & 25,986 \\
Teacher agreement & 89.2\% \\
\bottomrule
\end{tabular}

%% file: sections/experiments.tex
\section{Experiments}
\label{sec:exp}

We evaluate \ours around four research questions.
\textbf{RQ1}: Does \ours cut cost and latency on unseen workflows without lowering task success?
\textbf{RQ2}: How do the gate, executed states, and calibrated loss each contribute?
\textbf{RQ3}: What governs the savings?
\textbf{RQ4}: Does calibrated confidence select the right decisions to delegate?

\subsection{Experimental Setup}
\label{sec:exp-setup}

\paragraph{Benchmarks.}
To test cross-domain generalization without retraining, we evaluate on three unseen agent benchmarks spanning policy-constrained dialogue, partially observable embodied search, and noisy web navigation, with zero skill overlap with \data (Appendices~\ref{app:data} and~\ref{app:bench}):
(1)~\emph{SOPBench} \citep{li2025sopbench} evaluates multi-turn customer-service dialogues under strict procedural and database constraints across 268 held-out tasks, compiled deterministically from machine-readable SOPs;
(2)~\emph{ALFWorld} \citep{shridhar2021alfworld} tests multi-room object search and manipulation across 134 unseen text-based household tasks, compiled from SkillRL skill banks \citep{xia2026skillrl} into six task-family programs; and
(3)~\emph{WebShop} \citep{yao2022webshop} evaluates goal-directed shopping over a 1.18M-product web catalog across 500 user instructions (search formulation, attribute/price checks, and option selection within 15 steps), compiled from SkillRL skills into seven category programs.

\paragraph{Models and configurations.}
We pair a single 4B \sone trained exclusively on \data with DeepSeek-V4.1-Flash as the primary \stwo and GPT-6 Sol as a cross-family \stwo (decoding settings in Appendix~\ref{app:prereg}).
We compare three baselines—(i)~an uncompiled \emph{\stwo agent with skills} reading raw skill documents directly (the SOPBench reference agent; ReAct with SkillRL skills on ALFWorld and WebShop); (ii)~\emph{Compiled workflow (All-S2)}, where \stwo executes every E and G node; and (iii)~commercial \emph{Jev as \sone} (\texttt{jev-1.14.0}; \citealp{typesafe2026jev})—against two \ours variants: \emph{\gmain}, whose scalar threshold $\tau$ is calibrated on 50 unlabeled development runs per benchmark (\S\ref{sec:gate}), and cold-start \emph{\gzero}, which transfers a single threshold ($\tau=0.918$) with zero target-workflow runs.
When switching \stwo to GPT-6 Sol, \sone and \gzero remain completely unchanged while \gmain refits only $\tau$ on 50 unlabeled Sol development episodes; component ablations such as removing the gate (\emph{All-S1}) are analyzed in \S\ref{sec:rq2}.

\paragraph{Metrics and protocol.}
For task effectiveness, we report \emph{Pass} (\%) on SOPBench (tasks satisfying all final dialogue and database constraints), \emph{Success Rate (SR)} (\%) on ALFWorld, and both continuous \emph{Score} (0--100) and binary exact-match \emph{SR} (\%) on WebShop.
For efficiency, following \citet{zhu2026jevagentharness}, we report per-episode cost savings relative to the uncompiled \stwo agent with skills under the same \stwo (\emph{Sav.\,\%}; pricing \stwo and commercial Jev at official API list rates—\$0.042/1M input tokens and \$0 output for Jev—and our open-weight 4B \sone by self-hosted L4 GPU serving throughput at \$0.0045 per 1K single-pass decisions) alongside mean critical-path model-call time per episode (\emph{Time\,(s)}).
All thresholds were pre-registered before test runs, and each configuration shares an exact \stwo request cache with All-S2 so runs diverge strictly where \sone answers.

\subsection{RQ1: Main Results}
\label{sec:rq1}

\begin{table*}[t]
  \centering
  \input{tables/tab_online}
  \caption{End-to-end online evaluation on three unseen compiled benchmarks. Pass (\%), ALFWorld SR (\%), and WebShop Score / binary SR (\%) measure task effectiveness; Sav.\,\% reports episode cost savings relative to each block's uncompiled \stwo agent with skills (--: reference); Time\,(s) reports mean critical-path model-call seconds per episode; and Avg.\ reports the macro-mean primary-metric difference from All-S2 ($\Delta$\,pt across Pass, ALFWorld SR, and Score) and the mean Sav.\,\%. $^\dagger$Transferred $\tau$ \citep{typesafe2026jev}. Bold: block best.}
  \label{tab:online}
\end{table*}

\paragraph{Skill compilation and calibrated delegation provide complementary efficiency gains.}
Table~\ref{tab:online} separates two orthogonal levers behind end-to-end efficiency.
First, compiling unstructured skills into explicit D/E/G workflows (All-S2) removes mechanical tool orchestration from LLM turns and arrests prompt accumulation across dialogue history, preventing step-skipping, lifting macro-average effectiveness by nearly 18 points, and saving 76--86\% of the uncompiled agent's episode cost (81\% on average).
Second, offloading routine E decisions to single-pass \sone under \gmain raises the average saving to 89\% (86--91\% across benchmarks) and cuts critical-path model time by 70--92\% relative to the uncompiled agent (vs.\ 51--87\% for All-S2), while matching All-S2 reliability on average (+0.5 avg.\ pt).
In fact, single-pass local scoring suppresses verbose \stwo over-deliberation on straightforward binary relevance checks, improving WebShop score over All-S2 (68.1 vs.\ 66.9).
Compared against commercial Jev, self-hosted 4B \ours saves more of the uncompiled agent's cost (89\% vs.\ 84\% on average) at lower per-episode latency by avoiding remote API round-trips.
Even without target-workflow runs, cold-start \gzero stays level with All-S2 (+0.1 avg.\ pt) while saving 87\% of the uncompiled agent's cost on average.

\paragraph{A single \sone generalizes across \stwo model families without retraining.}
Replacing DeepSeek-V4.1-Flash with GPT-6 Sol reproduces the same two-tier pattern with the unchanged \sone weights (Table~\ref{tab:online}, bottom): compilation alone saves 71\% of single-system Sol's cost on average (56--86\%) at +16.8 avg.\ pt, and \gmain raises the saving to 82\% (75--91\%) while cutting model time by 63--84\% (vs.\ 40--74\% for All-S2).
Cold-start \gzero saves 81\% within 0.8 avg.\ pt of All-S2 without any Sol trajectories.
Because Sol has higher token pricing than Flash, fixed 4B inference accounts for an even smaller fraction of episode cost, underscoring that \data captures workflow-intrinsic decision semantics rather than teacher-specific priors.

\subsection{RQ2: Component Ablation Study}
\label{sec:rq2}

\begin{table}[t]
  \centering
  \input{tables/tab_ablation}
  \caption{Component ablations: (a)~online effectiveness and cost relative to the uncompiled \stwo agent with skills (in parentheses) across deferral configurations on DeepSeek-V4.1-Flash; (b)~three-seed decision accuracy (\%) across training state sources on the held-out container development set (Dev, 185 states) and benchmark dev splits (Table~\ref{tab:offline}); and (c)~single-seed offline accuracy, expected calibration error (ECE), and gated agreement (\%), averaged over Dev, SOP E, and ALF E.}
  \label{tab:ablation}
\end{table}

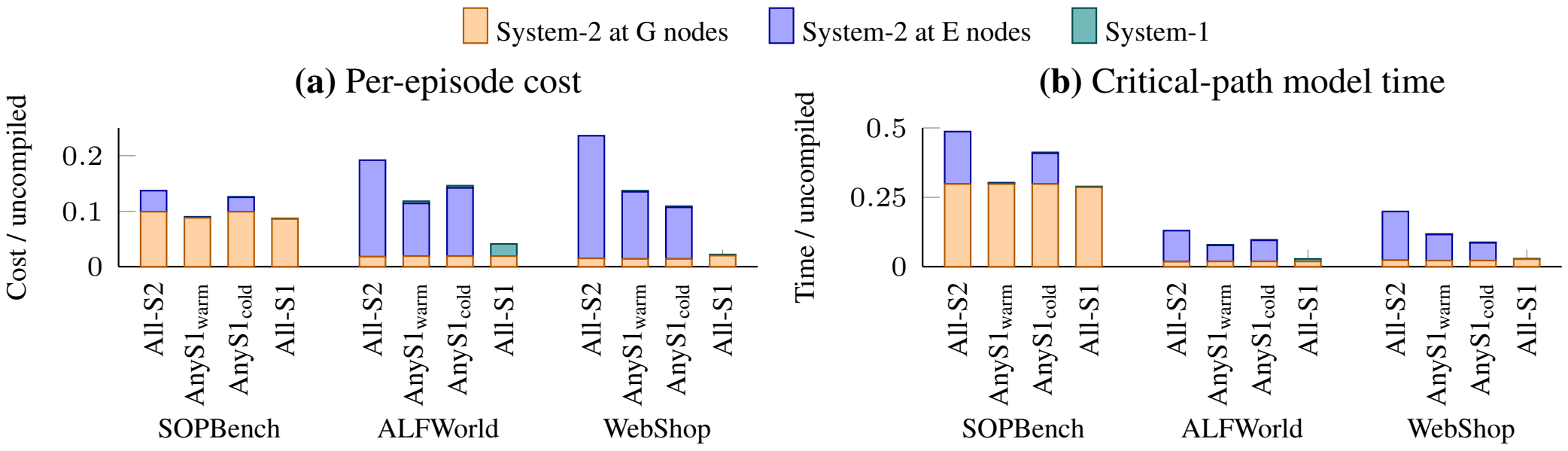
\begin{figure*}[t]
  \centering
  \input{figures/fig_cost}
  \caption{Efficiency relative to the uncompiled \stwo agent (DeepSeek-V4.1-Flash) via Eq.~\ref{eq:cost}: (a)~episode cost and (b)~critical-path model time ($21.8\text{--}31.7$\,ms \sone vs.\ $1.51\text{--}1.78$\,s \stwo E calls).}
  \label{fig:cost}
\end{figure*}

To isolate the role of each component, Table~\ref{tab:ablation} evaluates ablations along three dimensions: (a)~runtime deferral gating, (b)~training state source, and (c)~calibration objective (detailed setups in Appendix~\ref{app:ablation}).

\paragraph{Confidence gating prevents cascading search failures in long horizons.}
Comparing our calibrated gates (\gmain and \gzero) against ungated \emph{All-S1}---which delegates every E decision to \sone unconditionally---shows that short-horizon routing workflows remain intact without gating, whereas long-horizon ALFWorld exploration collapses ($96.3\% \to 40.3\%$).
Both \gmain and \gzero intercept uncertain open-ended exploration states via calibrated ranking rather than lower delegation volume—equal-rate random delegation multiplies \stwo disagreements more than fourfold (Appendix~\ref{app:gate}).

\paragraph{Container-executed states bridge the observation gap left by synthetic summaries.}
We compare \data against \emph{LLM-written synthetic} (\syn), which trains the same 4B architecture on LLM-drafted state summaries conditioned on mined decision points without container execution.
Training on \data's executed container states outperforms \syn across every split—most notably by +18.8 points on the held-out container development set (Dev) and +5.3 points on ALFWorld E—plausibly because idealized synthetic summaries mismatch terse, truncated terminal outputs.

\paragraph{Label smoothing prevents out-of-domain confidence saturation.}
Replacing Eq.~\ref{eq:loss} with unsmoothed \emph{hard-label cross-entropy} or \emph{CE + Brier} on the same \data training set yields nearly unchanged raw accuracy yet degrades out-of-domain calibration (Table~\ref{tab:ablation}c).
Because unsmoothed logits saturate near probability $1.0$ even on unfamiliar states, the dev-fitted threshold inflates toward $\tau=0.998$ and degrades cross-domain gated agreement (mean agreement drops from 97.7\% to 92.4--93.7\%), whereas label smoothing preserves uniform probability mass and a transferable confidence margin (Appendix~\ref{app:cal}).

\subsection{RQ3: What Governs the Savings?}
\label{sec:rq3}

\paragraph{Workflow E-node share governs both cost and latency ceilings.}
Normalizing episode cost and time to the uncompiled agent via Eq.~\ref{eq:cost} (Figure~\ref{fig:cost}) shows that post-compilation delegation targets the remaining E share: E nodes hold 17--22\% of uncompiled cost and 11--18\% of time on ALFWorld and WebShop, vs.\ 4\% and 19\% on SOPBench (G nodes: 10\% and 30\%).
Delegating nearly all SOPBench E nodes under \gmain approaches this bound, plus a residual G-node drop (9.9\% to 8.8\%).

\paragraph{Single-pass scoring converts delegation into critical-path speedup.}
Locally served \sone scores all options in $21.8\text{--}31.7$\,ms median (Appendix~\ref{app:latency}), $56\text{--}81\times$ faster than $1.51\text{--}1.78$\,s \stwo calls.
Under \gmain, \sone runtime takes 0.2--0.3\% of uncompiled episode time (Figure~\ref{fig:cost}b) while avoiding remote API round-trips (cf.\ $0.65\text{--}0.67$\,s for commercial Jev), cutting model time from 13--49\% (All-S2) to 8--30\%.

\subsection{RQ4: Does Confidence Select the Right Decisions?}
\label{sec:rq4}

\begin{figure}[t]
  \centering
  \input{figures/fig_opcurve}
  \caption{Operating curves on All-S2 test E decisions (DeepSeek-V4.1-Flash) vs.\ equal-rate random delegation (dashed); markers mark online thresholds.}
  \label{fig:opcurve}
\end{figure}
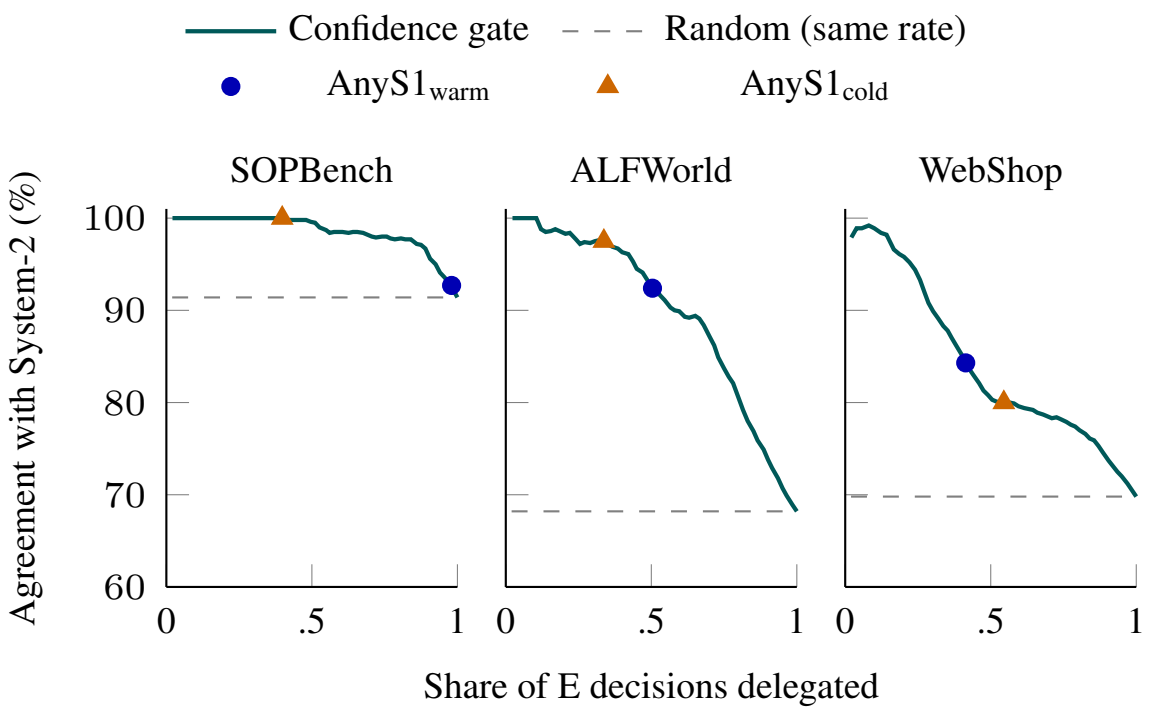

\paragraph{Confidence ranking, not delegation volume, keeps delegated decisions correct.}
Scoring All-S2 test E decisions and sweeping $\tau$ yields per-benchmark operating curves (Figure~\ref{fig:opcurve}).
At every coverage below 100\%, \sone's high-confidence picks match \stwo far better than random subsets of equal size, cutting AURC by $2\text{--}6\times$ (Table~\ref{tab:random}).
At \gmain's online threshold, ALFWorld delegates 50\% at 92.4\% agreement vs.\ 68.2\% random (3.8 vs.\ 16.0 disagreements per 100), WebShop delegates 41\% at 84.3\% vs.\ 69.8\%, and SOPBench delegates 98\% (91\% overall agreement).

\paragraph{A few development episodes place each workflow on its own operating point.}
Because curves differ across workflows, one threshold cannot serve all domains: \gzero's transferred $\tau$ is conservative on SOPBench (40\% delegated at 100\% agreement) yet over-delegates on WebShop (55\% at 80.0\%).
Target-workflow calibration avoids both using only 20 development episodes (matching full-fit delegation within 1 point on all benchmarks), whereas 5 episodes span 19--75\% on WebShop (10th--90th percentile).

%% file: tables/tab_online.tex
\footnotesize
\setlength{\tabcolsep}{2.4pt}
\begin{tabular}{@{}lcccccccccccc@{}}
\toprule
& \multicolumn{3}{c}{SOPBench} & \multicolumn{3}{c}{ALFWorld} & \multicolumn{4}{c}{WebShop} & \multicolumn{2}{c}{Avg.} \\
\cmidrule(lr){2-4}\cmidrule(lr){5-7}\cmidrule(lr){8-11}\cmidrule(lr){12-13}
Configuration & Pass & Sav.\,\% & Time\,(s) & SR & Sav.\,\% & Time\,(s) & Score & SR & Sav.\,\% & Time\,(s) & $\Delta$\,pt & Sav.\,\% \\
\midrule
\multicolumn{13}{@{}l}{\textit{\stwo: DeepSeek-V4.1-Flash}} \\
\stwo agent with skills & 55.2 & -- & 51.4 & 82.8 & -- & 163 & 54.9 & 34.4 & -- & 146 & $-$17.8 & -- \\
Compiled workflow (All-S2) & 83.2 & 86 & 25.0 & \textbf{96.3} & 81 & 21.3 & 66.9 & 38.0 & 76 & 29.0 & 0.0 & 81 \\
Jev as \sone$^\dagger$ & 84.0 & 87 & 23.9 & \textbf{96.3} & 84 & 20.5 & 67.1 & 38.6 & 83 & 17.3 & +0.3 & 84 \\
\gmain & \textbf{84.3} & \textbf{91} & \textbf{15.6} & 95.5 & \textbf{88} & \textbf{12.9} & \textbf{68.1} & \textbf{38.8} & 86 & 17.1 & \textbf{+0.5} & \textbf{89} \\
\gzero & 83.2 & 87 & 21.2 & 95.5 & 85 & 15.7 & 67.9 & 38.0 & \textbf{89} & \textbf{12.7} & +0.1 & 87 \\
\midrule
\multicolumn{13}{@{}l}{\textit{\stwo: GPT-6 Sol}} \\
\stwo agent with skills & 61.2 & -- & 207.6 & 88.8 & -- & 413.5 & 48.5 & 33.6 & -- & 317.3 & $-$16.8 & -- \\
Compiled workflow (All-S2) & 82.1 & 86 & 124 & \textbf{96.3} & 56 & 193 & \textbf{70.6} & \textbf{41.8} & 72 & 81.4 & 0.0 & 71 \\
\gmain & \textbf{82.5} & \textbf{91} & \textbf{75.8} & 95.5 & \textbf{75} & \textbf{101} & 70.4 & 40.0 & 80 & 49.4 & $-$0.2 & \textbf{82} \\
\gzero & \textbf{82.5} & 88 & 107 & 94.0 & 69 & 134 & 70.1 & 40.4 & \textbf{86} & \textbf{33.3} & $-$0.8 & 81 \\
\bottomrule
\end{tabular}

%% file: tables/tab_ablation.tex
\footnotesize
\setlength{\tabcolsep}{3.2pt}
\resizebox{\columnwidth}{!}{%
\begin{tabular}{@{}lccc@{}}
\toprule
\textit{(a) Runtime deferral gate} & \textbf{SOP} (Cost) & \textbf{ALF} (Cost) & \textbf{WS} (Cost) \\
\midrule
\gmain & \textbf{84.3} (0.09$\times$) & \textbf{95.5} (0.12$\times$) & \textbf{68.1} (0.14$\times$) \\
\gzero & 83.2 (0.13$\times$) & \textbf{95.5} (0.15$\times$) & 67.9 (0.11$\times$) \\
All-S1 & \textbf{84.3} (0.09$\times$) & 40.3 (0.04$\times$) & 66.3 (0.02$\times$) \\
\midrule
\textit{(b) Training state source} & \textbf{Dev} & \textbf{SOP E} & \textbf{ALF E} \\
\midrule
\data & \textbf{83.8} & \textbf{93.5} & \textbf{68.9} \\
LLM-written synthetic & 65.0 & 91.7 & 63.6 \\
\midrule
\textit{(c) Calibration objective} & \textbf{Mean Acc.} & \textbf{Mean ECE $\downarrow$} & \textbf{Mean Agr.} \\
\midrule
Label smoothing & \textbf{82.4} & \textbf{0.094} & \textbf{97.7} \\
Hard-label CE & 82.3 & 0.146 & 92.4 \\
CE + Brier & 81.9 & 0.148 & 93.7 \\
\bottomrule
\end{tabular}%
}

%% file: figures/fig_cost.tex
\begin{tikzpicture}
\begin{axis}[
  name=costax,
  ybar stacked, bar width=6.5pt, width=0.45\textwidth, height=2.8cm,
  ymin=0, ymax=0.25, ytick={0,0.1,0.2},
  ylabel={Cost / uncompiled}, ylabel style={font=\scriptsize},
  title={\textbf{(a)} Per-episode cost}, title style={font=\footnotesize, yshift=-3pt},
  xtick={0,1,2,3,5,6,7,8,10,11,12,13}, xticklabels={All-S2,\gmain,\gzero,All-S1,All-S2,\gmain,\gzero,All-S1,All-S2,\gmain,\gzero,All-S1},
  xticklabel style={font=\scriptsize, rotate=90, anchor=east},
  yticklabel style={font=\scriptsize}, xmin=-0.8, xmax=13.8,
  legend style={font=\scriptsize, at={(1.13,1.0)}, anchor=south, yshift=16pt, legend columns=3, draw=none, /tikz/every even column/.append style={column sep=8pt}},
  legend cell align=left, axis x line*=bottom, axis y line*=left, clip=false,
  enlarge x limits=false,
]
\addplot+[fill=orange!35, draw=orange!70!black] coordinates {(0,0.099) (1,0.088) (2,0.099) (3,0.086) (5,0.018) (6,0.019) (7,0.019) (8,0.019) (10,0.015) (11,0.014) (12,0.014) (13,0.020)};
\addlegendentry{\stwo at G nodes}
\addplot+[fill=blue!35, draw=blue!60!black] coordinates {(0,0.038) (1,0.001) (2,0.026) (3,0.000) (5,0.174) (6,0.095) (7,0.123) (8,0.000) (10,0.221) (11,0.121) (12,0.093) (13,0.000)};
\addlegendentry{\stwo at E nodes}
\addplot+[fill=teal!55, draw=teal!60!black] coordinates {(0,0.000) (1,0.001) (2,0.001) (3,0.001) (5,0.000) (6,0.004) (7,0.004) (8,0.022) (10,0.000) (11,0.002) (12,0.002) (13,0.002)};
\addlegendentry{\sone}
\node[font=\scriptsize, anchor=north, yshift=-34pt] at (axis cs:1.5,0) {SOPBench};
\node[font=\scriptsize, anchor=north, yshift=-34pt] at (axis cs:6.5,0) {ALFWorld};
\node[font=\scriptsize, anchor=north, yshift=-34pt] at (axis cs:11.5,0) {WebShop};
\end{axis}
\begin{axis}[
  name=timeax, at={(costax.south east)}, anchor=south west, xshift=1.45cm,
  ybar stacked, bar width=6.5pt, width=0.45\textwidth, height=2.8cm,
  ymin=0, ymax=0.5, ytick={0,0.25,0.5},
  ylabel={Time / uncompiled}, ylabel style={font=\scriptsize},
  title={\textbf{(b)} Critical-path model time}, title style={font=\footnotesize, yshift=-3pt},
  xtick={0,1,2,3,5,6,7,8,10,11,12,13}, xticklabels={All-S2,\gmain,\gzero,All-S1,All-S2,\gmain,\gzero,All-S1,All-S2,\gmain,\gzero,All-S1},
  xticklabel style={font=\scriptsize, rotate=90, anchor=east},
  yticklabel style={font=\scriptsize}, xmin=-0.8, xmax=13.8,
  axis x line*=bottom, axis y line*=left, clip=false,
  enlarge x limits=false,
]
\addplot+[fill=orange!35, draw=orange!70!black] coordinates {(0,0.298) (1,0.298) (2,0.298) (3,0.286) (5,0.018) (6,0.019) (7,0.019) (8,0.019) (10,0.023) (11,0.022) (12,0.022) (13,0.027)};
\addplot+[fill=blue!35, draw=blue!60!black] coordinates {(0,0.189) (1,0.002) (2,0.111) (3,0.000) (5,0.112) (6,0.058) (7,0.076) (8,0.000) (10,0.176) (11,0.094) (12,0.064) (13,0.000)};
\addplot+[fill=teal!55, draw=teal!60!black] coordinates {(0,0.000) (1,0.003) (2,0.003) (3,0.003) (5,0.000) (6,0.002) (7,0.002) (8,0.009) (10,0.000) (11,0.002) (12,0.002) (13,0.002)};
\node[font=\scriptsize, anchor=north, yshift=-34pt] at (axis cs:1.5,0) {SOPBench};
\node[font=\scriptsize, anchor=north, yshift=-34pt] at (axis cs:6.5,0) {ALFWorld};
\node[font=\scriptsize, anchor=north, yshift=-34pt] at (axis cs:11.5,0) {WebShop};
\end{axis}
\end{tikzpicture}

%% file: figures/fig_opcurve.tex
\begin{tikzpicture}
\begin{axis}[name=p0, width=0.25\columnwidth, height=2.5cm, scale only axis, xmin=0, xmax=1, ymin=60, ymax=101, xtick={0,0.5,1}, xticklabels={0,.5,1}, ytick={60,70,80,90,100}, tick label style={font=\scriptsize}, label style={font=\scriptsize}, legend style={font=\scriptsize, at={(0,1.22)}, anchor=south west, legend columns=2, draw=none, /tikz/every even column/.append style={column sep=4pt}}, ylabel={Agreement with \stwo (\%)}, axis x line*=bottom, axis y line*=left, clip=false]
\addplot[teal!70!black, thick] coordinates {(0.021,100.0) (0.039,100.0) (0.056,100.0) (0.076,100.0) (0.090,100.0) (0.106,100.0) (0.126,100.0) (0.141,100.0) (0.155,100.0) (0.170,100.0) (0.191,100.0) (0.206,100.0) (0.220,100.0) (0.241,100.0) (0.257,100.0) (0.270,100.0) (0.285,100.0) (0.306,100.0) (0.321,100.0) (0.335,100.0) (0.350,100.0) (0.371,100.0) (0.385,100.0) (0.398,100.0) (0.415,99.8) (0.430,99.8) (0.446,99.8) (0.461,99.8) (0.480,99.8) (0.496,99.6) (0.511,99.5) (0.526,99.0) (0.549,98.7) (0.561,98.4) (0.576,98.5) (0.605,98.5) (0.626,98.4) (0.641,98.5) (0.656,98.5) (0.676,98.4) (0.691,98.2) (0.706,98.0) (0.720,97.9) (0.741,98.0) (0.757,98.0) (0.770,97.8) (0.785,97.7) (0.806,97.8) (0.821,97.7) (0.840,97.7) (0.860,97.2) (0.875,97.1) (0.890,96.7) (0.906,95.6) (0.925,95.0) (0.940,94.1) (0.955,93.6) (0.971,93.0) (0.986,92.2) (1.000,91.4)};
\addplot[gray, dashed] coordinates {(0.02,91.4) (1,91.4)};
\addplot[only marks, mark=*, mark size=1.6pt, blue!70!black] coordinates {(0.980,92.7)};
\addplot[only marks, mark=triangle*, mark size=2.1pt, orange!80!black] coordinates {(0.398,100.0)};
\node[font=\scriptsize, anchor=base] at (rel axis cs:0.5,1.07) {SOPBench};
\legend{{Confidence gate},{Random (same rate)},\gmain,\gzero}
\end{axis}
\begin{axis}[name=p1, at={(p0.south east)}, anchor=south west, xshift=0.32cm, width=0.25\columnwidth, height=2.5cm, scale only axis, xmin=0, xmax=1, ymin=60, ymax=101, xtick={0,0.5,1}, xticklabels={0,.5,1}, ytick={60,70,80,90,100}, tick label style={font=\scriptsize}, label style={font=\scriptsize}, yticklabels={}, xlabel={Share of E decisions delegated}, axis x line*=bottom, axis y line*=left, clip=false]
\addplot[teal!70!black, thick] coordinates {(0.023,100.0) (0.035,100.0) (0.055,100.0) (0.071,100.0) (0.085,100.0) (0.105,100.0) (0.121,98.8) (0.137,98.5) (0.155,98.6) (0.170,98.8) (0.191,98.5) (0.205,98.3) (0.220,98.4) (0.241,97.7) (0.255,97.2) (0.270,97.4) (0.290,97.3) (0.306,97.5) (0.320,97.6) (0.337,97.5) (0.350,97.2) (0.366,96.9) (0.385,96.7) (0.400,96.3) (0.421,96.1) (0.437,95.3) (0.450,94.5) (0.470,94.1) (0.486,93.3) (0.500,92.5) (0.516,92.2) (0.531,91.6) (0.546,91.1) (0.566,90.3) (0.580,90.0) (0.596,89.9) (0.615,89.3) (0.630,89.2) (0.651,89.4) (0.665,89.1) (0.680,88.4) (0.701,87.1) (0.716,86.2) (0.730,84.9) (0.750,83.7) (0.766,82.8) (0.781,82.1) (0.801,80.4) (0.815,79.2) (0.831,78.0) (0.851,76.9) (0.865,75.9) (0.886,74.9) (0.901,73.8) (0.915,72.9) (0.935,71.8) (0.951,70.7) (0.965,69.9) (0.985,68.9) (1.000,68.2)};
\addplot[gray, dashed] coordinates {(0.02,68.2) (1,68.2)};
\addplot[only marks, mark=*, mark size=1.6pt, blue!70!black] coordinates {(0.504,92.4)};
\addplot[only marks, mark=triangle*, mark size=2.1pt, orange!80!black] coordinates {(0.337,97.5)};
\node[font=\scriptsize, anchor=base] at (rel axis cs:0.5,1.07) {ALFWorld};
\end{axis}
\begin{axis}[name=p2, at={(p1.south east)}, anchor=south west, xshift=0.32cm, width=0.25\columnwidth, height=2.5cm, scale only axis, xmin=0, xmax=1, ymin=60, ymax=101, xtick={0,0.5,1}, xticklabels={0,.5,1}, ytick={60,70,80,90,100}, tick label style={font=\scriptsize}, label style={font=\scriptsize}, yticklabels={}, axis x line*=bottom, axis y line*=left, clip=false]
\addplot[teal!70!black, thick] coordinates {(0.021,97.9) (0.039,98.9) (0.057,98.9) (0.081,99.2) (0.100,98.9) (0.123,98.4) (0.142,98.2) (0.167,96.6) (0.186,96.1) (0.202,95.8) (0.221,95.2) (0.240,94.4) (0.257,93.3) (0.273,91.9) (0.286,90.8) (0.302,89.9) (0.320,89.1) (0.337,88.3) (0.352,87.8) (0.366,87.0) (0.381,86.2) (0.400,85.2) (0.414,84.3) (0.425,83.7) (0.440,83.0) (0.460,82.1) (0.475,81.3) (0.491,80.8) (0.505,80.3) (0.520,80.1) (0.540,80.0) (0.550,80.0) (0.565,80.0) (0.580,79.9) (0.595,79.6) (0.615,79.4) (0.630,79.3) (0.645,79.2) (0.660,78.9) (0.680,78.7) (0.695,78.5) (0.710,78.3) (0.726,78.4) (0.740,78.2) (0.760,77.9) (0.775,77.6) (0.790,77.4) (0.805,77.0) (0.825,76.6) (0.840,76.1) (0.855,75.9) (0.870,75.3) (0.885,74.6) (0.905,73.7) (0.920,73.1) (0.935,72.5) (0.950,72.0) (0.970,71.2) (0.985,70.5) (1.000,69.8)};
\addplot[gray, dashed] coordinates {(0.02,69.8) (1,69.8)};
\addplot[only marks, mark=*, mark size=1.6pt, blue!70!black] coordinates {(0.414,84.3)};
\addplot[only marks, mark=triangle*, mark size=2.1pt, orange!80!black] coordinates {(0.545,80.0)};
\node[font=\scriptsize, anchor=base] at (rel axis cs:0.5,1.07) {WebShop};
\end{axis}
\end{tikzpicture}

%% file: sections/conclusion.tex
\section{Conclusion}
\label{sec:conclusion}

We presented \ours, turning on-demand procedural skills into a shared source of structured decision points without custom per-workflow \sone engineering.
By mining grounded enumerable judgments across public skills, collecting container-executed states, and training one label-smoothed pointer \sone, \ours resolves routine steps and calibrates deferral to \stwo in one forward pass.
Experiments across unseen benchmarks and distinct \stwo model families show that both unlabeled calibration and zero-shot threshold transfer preserve all-\stwo reliability at substantially lower cost and critical-path latency.

%% file: sections/limitations.tex
\section*{Limitations}

While \gzero transfers with zero target-workflow executions, achieving the largest cost reductions under \gmain uses about 50 development episodes per workflow in which \stwo answers every E node, and large episode-level cost savings rely on compiling procedural skills into explicit programs; adapting deferral thresholds online from deferred \stwo calls and automatically compiling open-ended repository skills are natural next steps.

%% file: sections/appendix.tex
\section{\data Construction, Prompts, and Worked Example}
\label{app:data}

\paragraph{Source curation and split isolation.}
We curate procedural skills from SkillMD-138K \citep{skillmd138k} according to four criteria:
(i)~\emph{procedural structure and licensing}, retaining multi-step English procedural documents released under permissive open-source SPDX licenses while excluding bulk aggregator repositories;
(ii)~\emph{diversity and deduplication}, applying MinHash near-duplicate removal and capping at most three skills per repository so no single codebase dominates;
(iii)~\emph{split isolation and benchmark decontamination}, partitioning repositories into disjoint training and held-out sets beforehand and excluding any skill that references downstream evaluation or public containerized agent benchmarks (SOPBench, ALFWorld, WebShop, SkillRL, or $\tau$-bench); and
(iv)~\emph{sandbox executability}, verifying that each skill can run end-to-end inside an offline Linux container without external accounts or live human input.
Stratified sampling across 16 functional categories yields the 1,072 public training skills in \data.

\paragraph{Decision-point filtering and task synthesis.}
Across three extraction passes per skill (Figure~\ref{fig:prompts}, Prompt~1), an LLM merger retains E nodes with support $\sigma(p)\ge 2$ that express non-preference judgments, are answerable from local text context, and include at least two candidate options alongside \none.
For skills with at least three retained E nodes, DeepSeek-V4.1-Flash generates multi-stage Linux tasks paired with cross-category distractors, a hidden verifier, and a reference solution that passes inside an isolated container (Prompt~2).

\paragraph{Execution labeling and dataset summary.}
OpenHands agents \citep{wang2024openhands} execute the verified tasks, a trajectory locator copies verbatim pre-decision observations (user messages and tool outputs, excluding agent reasoning), and two order-reversed teacher passes assign consensus labels (Prompt~3).
Overall, 2,846 verified tasks across 1,072 skills yield 17,865 mined decision points and 25,986 consensus examples (89.2\% teacher agreement; median 3 options besides \none; 6.7\% labeled \none).
Applying the same labeling protocol to disjoint Apache-2.0 containerized trajectories gives the 153 calibration states (\gzero) and 185 offline development states (\emph{Dev}, Appendix~\ref{app:offline}).

\paragraph{Prompt templates and synthesized example.}
Figure~\ref{fig:prompts} shows the prompt templates for the three offline LLM stages used to build \data, and Figure~\ref{fig:example} illustrates an authentic end-to-end synthesized decision record from \data, showing how a mined E-node template $(q,O,d)$ is grounded into a container-executed observation state $s$, serialized via Eq.~\ref{eq:input}, and labeled by order-reversed dual-teacher consensus.

\begin{figure*}[t]
  \centering
  \begin{promptbox}{Prompt 1: Static Decision-Point Extraction (\S\ref{sec:mining}, 3 independent passes + merge)}
    \prole{System:} You analyze agent skills (procedural documents an LLM agent loads to perform a task) and extract the runtime decision points an agent encounters when following \texttt{SKILL.md} and bundled documentation. Output JSON only.

    \prole{Rules:} Classify each decision point in priority order as \textbf{D} (deterministic check fully answerable from structured state or code), \textbf{E} (enumerable judgment reading unstructured text or observations over a finite option set, always allowing \textit{none of the above}), or \textbf{G} (open-ended generation). For each \textbf{E} node, specify its judgment kind (\texttt{literal} | \texttt{inferential} | \texttt{planning} | \texttt{preference}), list candidate options $O$ with one-sentence definitions $d(o)$, and copy a verbatim \texttt{evidence} quote from the skill text.

    \prole{User:} Inspect skill \pvar{name} (\texttt{SKILL.md} and referenced docs \pvar{extra\_docs}) with bundled file list \pvar{files}, and return JSON \texttt{\{"skill": ..., "domain": ..., "decision\_points": [...]\}}.
  \end{promptbox}

  \vspace{4pt}
  \begin{promptbox}{Prompt 2: Multi-Stage Container Task \& Verifier Synthesis (\S\ref{sec:states}, container-checked)}
    \prole{User:} Design a realistic offline Linux container task for target skill \pvar{skill} containing E decision points \pvar{points}, alongside cross-category distractor skills \pvar{distractors}.

    \prole{Requirements:} (1) Require at least three dependent stages where facts settling the target E nodes are spread across workspace files or revealed by intermediate computation, including at least one stale or conflicting source to reconcile; never quote decision questions or name the target skill. (2) Return a user \texttt{instruction}, workspace \texttt{files}/\texttt{setup.py}, a deterministic \texttt{verify.py} (must fail on initial files), and a hidden reference \texttt{solution.py} (must pass \texttt{verify.py} inside a network-isolated container).
  \end{promptbox}

  \vspace{4pt}
  \begin{promptbox}{Prompt 3: Trajectory Locator \& Order-Reversed Dual-Teacher Labeling (\S\ref{sec:states})}
    \prole{Locator:} Given numbered execution transcript \pvar{transcript} and mined E decision points \pvar{points}, identify every step where the agent faced each decision and copy verbatim the preceding observation passages (\texttt{User request}, \texttt{Tool output}; exclude agent reasoning) as state $s$. Return JSON \texttt{\{"points": [\{"id": ..., "occurs": ..., "step": ..., "passages": [...], "later": [...]\}]\}}.

    \tcbsep
    \prole{Teacher (Pass 1 \& Pass 2):} \textit{``You are a decision component inside an agent following the skill \pvar{skill}. Context: \texttt{<<<}\pvar{state}\texttt{>>>} Question: \pvar{question} Options: \pvar{options\_with\_definitions}. Answer with JSON only: \texttt{\{"answer": "<one option exactly as written>"\}}.''} (Pass~2 reverses candidate option order while keeping \textit{none of the above} last; retain iff $y^{1}=y^{2}$.)
  \end{promptbox}
  \caption{Abridged prompt templates for the three offline LLM stages in \data construction.}
  \label{fig:prompts}
\end{figure*}

\begin{figure*}[t]
  \centering
  \begin{examplebox}{Worked Example from \data (Skill: \texttt{brand-yml} $\mid$ Domain: Web \& Reporting $\mid$ Support $\sigma=3$)}
    \erole{1. Mined E-Node Template $(q, O, d)$:}
    \par\noindent\hspace*{0.8em}$q$: \textit{Should the brand file use the standard auto-discovered \texttt{\_brand.yml} name/location, or a custom name/location that requires explicit paths?}
    \par\noindent\hspace*{0.8em}$\bullet$ $o_1$ = \texttt{standard \_brand.yml}: \textit{File can be named \texttt{\_brand.yml} at the project/app root or standard \texttt{\_brand/} directory so Shiny or Quarto discovers it automatically.}
    \par\noindent\hspace*{0.8em}$\bullet$ $o_2$ = \texttt{custom name/location with explicit path}: \textit{Brand uses a custom filename or nonstandard shared directory, so theme calls or frontmatter must pass an explicit path.}
    \par\noindent\hspace*{0.8em}$\bullet$ $o_3$ = \texttt{none of the above}: \textit{No listed option applies.}

    \tcbsep
    \erole{2. Container-Executed Runtime State $s$ (verbatim pre-decision observations):}
    \par\noindent\hspace*{0.8em}\texttt{User request:} Write a single canonical brand file at \texttt{brand/meridian\_brand.yml} \dots\ Update the R app's theme call in \texttt{tools/shiny-r/app.R} so it loads the brand file via an explicit path (the R app lives outside the project root) \dots
    \par\noindent\hspace*{0.8em}\texttt{Tool output (cat notes.md):} This app is built and deployed from \texttt{tools/shiny-r/} but the shared brand file lives under \texttt{brand/}, which is outside this directory. \texttt{bslib} is pinned to 0.9.x; supports \texttt{bs\_theme(brand = ...)} with either a boolean or a path string.

    \tcbsep
    \erole{3. Serialized \sone Input $\tilde{x}$ \& Consensus Label $y$:}
    \par\noindent\hspace*{0.8em}$\tilde{x} = \big[\langle\mathrm{s}\rangle, s;\, \langle\mathrm{q}\rangle, q;\, \langle\mathrm{opt}\rangle, o_1{:}\,d(o_1), \langle/\mathrm{opt}\rangle;\, \langle\mathrm{opt}\rangle, o_2{:}\,d(o_2), \langle/\mathrm{opt}\rangle;\, \langle\mathrm{opt}\rangle, o_3{:}\,d(o_3), \langle/\mathrm{opt}\rangle;\, \langle\mathrm{dec}\rangle\big]$ (Eq.~\ref{eq:input})
    \par\noindent\hspace*{0.8em}Forward teacher $y^{1}$ = Reversed teacher $y^{2}$ = \textbf{\textcolor{exampleframe}{\texttt{custom name/location with explicit path}}} (consensus label $y$).
  \end{examplebox}
  \caption{End-to-end synthesized decision record from \data (\texttt{brand-yml}). Container execution grounds the mined E-node template $(q,O,d)$ in multi-file user requests and tool outputs before order-reversed dual-teacher consensus labeling.}
  \label{fig:example}
\end{figure*}

\section{Implementation, Training, and Evaluation Protocols}
\label{app:training}
\label{app:bench}
\label{app:prereg}
\label{app:latency}
\label{app:ablation}

\paragraph{Architecture and training hyperparameters.}
We build \sone on Qwen3.5-4B-Base \citep{qwen2026qwen35} using rank-16 LoRA adapters \citep{hu2022lora} ($\alpha=32$) across attention, MLP, and Gated DeltaNet projections.
Above the pointer projection ($d_p=256$), we attach a permutation-equivariant option-interaction head consisting of a single position-free 4-head bidirectional self-attention layer and a $2\times$ MLP over the $\langle\mathrm{dec}\rangle$ query vector and all $\langle/\mathrm{opt}\rangle$ option vectors, with zero-initialized residual projections so initial logits match a standard pointer head.
We train for 1 epoch in bf16 with learning rate $10^{-4}$ and label smoothing $\epsilon=0.1$ (Eq.~\ref{eq:loss}).

\paragraph{Benchmark protocols, compilation, and decoding settings.}
ALFWorld (134 unseen test games) and WebShop (500 test instructions, 15-step limit, 1.18M-product catalog) follow standard benchmark splits; SOPBench evaluates on 268 held-out test tasks disjoint from development tasks used for compilation and calibration.
Across both \stwo backbones, decoding uses temperature 0 with reasoning enabled on G nodes and free-form agent turns and disabled on E nodes, and episode costs follow official API list rates.
On SOPBench, a deterministic rule compiler translates machine-readable SOPs, tool schemas, and constraint predicates into D nodes, E nodes, and G nodes.
On ALFWorld and WebShop, an LLM compiler translates the SkillRL skill bank \citep{xia2026skillrl} into one structured D/E/G program per task family (6 for ALFWorld, 7 product categories for WebShop; Figure~\ref{fig:compile-example}).

\paragraph{Latency measurement protocol.}
For Figure~\ref{fig:cost}b and the RQ3 latency comparison, \sone per-decision inference latency is benchmarked on a single NVIDIA H100 GPU over development E decisions across SOPBench ($n=207$, median 21.8\,ms), ALFWorld ($n=764$, 23.9\,ms), and WebShop ($n=401$, 31.7\,ms).
Per-decision \stwo latencies (DeepSeek-V4.1-Flash E calls under All-S2: median 1.78\,s on SOPBench, 1.51\,s on ALFWorld, and 1.77\,s on WebShop) and commercial Jev latencies (0.65\,s on ALFWorld, 0.67\,s on WebShop) are measured from online API request logs; episode critical-path model times in Figure~\ref{fig:cost}b sum logged \stwo calls and every \sone scoring pass (including passes that defer to \stwo) along each trajectory.

\paragraph{Ablation experimental setups.}
Each block of Table~\ref{tab:ablation} varies one factor under fixed pipeline settings:
(a)~\emph{Runtime deferral gate} compares \gmain ($\tau$ calibrated on 50 unlabeled development runs targeting $\ge 95\%$ \stwo agreement), \gzero (fixed $\tau=0.918$ transferred from the 153-state calibration split), and \emph{All-S1} ($\tau=0$) on DeepSeek-V4.1-Flash using shared All-S2 request caches.
(b)~\emph{Training state source} compares \data against \emph{LLM-written synthetic} (\syn, three seeds), trained under the identical Qwen3.5-4B-Base architecture and hyperparameters on 84,481 LLM-authored textual summaries generated from mined decision points ($3.3\times$ the 25,986 \data examples, so this block varies corpus size along with state source).
(c)~\emph{Calibration objective} trains the same model on 100\% \data while varying only the loss among label smoothing ($\epsilon=0.1$), hard-label cross-entropy ($\epsilon=0$), and cross-entropy plus a $1.0\times$ Brier penalty.

\begin{figure*}[t]
  \centering
  \begin{examplebox}{Skill Compilation Example: Raw SkillRL Skill Bank $\rightarrow$ Typed D/E/G Workflow (\texttt{WebShop/apparel})}
    \erole{1. Source Skill Bank Excerpts} (\texttt{claude\_style\_skills.json}, SkillRL \texttt{apparel} slice):
    \begin{itemize}\setlength{\itemsep}{1.5pt}\setlength{\parskip}{0pt}\setlength{\parsep}{0pt}
      \item \texttt{gen\_001} (\emph{Prioritize Core Keywords}) \& \texttt{app\_001} (\emph{Focus Key Query}): Include only product type + must-have attributes (gender, garment, fabric, sleeve/fit); drop minor/redundant terms before searching.
      \item \texttt{gen\_003} (\emph{Scan Before You Click}) \& \texttt{err\_003}: Read titles, thumbnails, and prices in results to check category metadata before opening a product page.
      \item \texttt{gen\_004}, \texttt{app\_002} (\emph{Validate Core Match}) \& \texttt{app\_004} (\emph{Check Variant Price}): Confirm budget and garment-type match immediately upon opening; leave at once if violated.
      \item \texttt{gen\_005} \& \texttt{app\_003} (\emph{Set Variants First}): Select required size and color options before re-checking variant price and purchasing.
    \end{itemize}
    \tcbsep
    \erole{2. Compiled Typed D/E/G Program} (\texttt{apparel/program.json}; skills linked per node):
    \begin{itemize}\setlength{\itemsep}{1.5pt}\setlength{\parskip}{0pt}\setlength{\parsep}{0pt}
      \item \textbf{[\texttt{G}] \texttt{s0} (\texttt{gen\_query})} $\leftarrow$ \texttt{[gen\_001, app\_001, err\_001]}: Write concise query \texttt{\{query\}} (product type + core attributes). $\rightarrow$ \textbf{[\texttt{D}] \texttt{q1}} \texttt{search[\{query\}]} $\rightarrow$ \textbf{[\texttt{D}] \texttt{inc1}} (\texttt{searches += 1}) $\rightarrow$ \textbf{[\texttt{D}] \texttt{r0}} (\texttt{branch unvisited\_left}).
      \item \textbf{[\texttt{E}] \texttt{pick} (\texttt{choose product\_pick})} $\leftarrow$ \texttt{[gen\_003, app\_002, err\_003]}: Score \texttt{unvisited\_results} $\cup$ \{\texttt{none}\}; bind \texttt{asin} ($\rightarrow$ \texttt{open}) or branch \texttt{on\_none} $\rightarrow$ \texttt{nopage}.
      \item \textbf{[\texttt{D}] \texttt{open} $\rightarrow$ \texttt{item1} $\rightarrow$ \texttt{item2}} $\leftarrow$ \texttt{[gen\_004, app\_004]}: Execute \texttt{click[\{asin\}]}, verify item page, run deterministic budget check \texttt{price\_ok} (\texttt{else} $\rightarrow$ \textbf{[\texttt{D}] \texttt{back}:} \texttt{click[< prev]}).
      \item \textbf{[\texttt{E}] \texttt{match} (\texttt{ask\_yes\_no core\_match})} $\leftarrow$ \texttt{[gen\_004, app\_002, err\_003]}: Check garment type, gender, and primary attributes (\texttt{on\_yes} $\rightarrow$ \texttt{variants}; \texttt{on\_no} $\rightarrow$ \texttt{back}).
      \item \textbf{[\texttt{D}] \texttt{variants} loop}: \textbf{[\texttt{E}] \texttt{vg} (\texttt{choose option\_value})} $\rightarrow$ \textbf{[\texttt{D}] \texttt{sel}} \texttt{click[\{value\}]}, or \texttt{on\_none} $\rightarrow$ \textbf{[\texttt{E}] \texttt{vgnone}} ($\texttt{yes}\to\texttt{back}$; $\texttt{no}\to\texttt{vgnext}$).
      \item \textbf{[\texttt{D}] \texttt{pricecheck} $\rightarrow$ \texttt{buy}} $\leftarrow$ \texttt{[app\_004, gen\_013]}: Deterministic variant price check; if \texttt{price\_ok}, execute \texttt{click[buy now]} (else \texttt{back}). Bounded search refinement (\textbf{[\texttt{G}] \texttt{s2}} $\rightarrow$ \textbf{[\texttt{D}] \texttt{q2}}) triggers only if \texttt{searches < 3} and \texttt{steps\_left >= 6}.
    \end{itemize}
  \end{examplebox}
  \caption{Skill compilation example from the WebShop \texttt{apparel} family. Unconstrained prose heuristics in SkillRL compile into a verifiable control graph separating deterministic code branches (\textbf{D}), closed-set judgments (\textbf{E}), and localized open generation (\textbf{G}).}
  \label{fig:compile-example}
\end{figure*}

\section{Additional Offline, Calibration, and Online Analyses}
\label{app:offline}
\label{app:cal}
\label{app:gate}
\label{app:online}

\paragraph{Full offline baseline comparison.}
Expanding Table~\ref{tab:ablation}(b), Table~\ref{tab:offline} evaluates \sone (three seeds) across our offline development sets (\emph{Dev}, \emph{SOPBench E}, \emph{ALFWorld E}, and its \emph{search} subset) against zero-shot typed decision models (Kev-4B, AgentDecision-4B, commercial Jev), zero-shot Qwen3.5-4B/27B, and \syn.
On held-out \emph{Dev}, \sone (83.8\%) outperforms \syn (65.0\%) and commercial Jev (78.4\%), while the option-interaction head improves over a standard pointer head (82.7\%) by $+1.1$ points; on \emph{ALFWorld search}, the stronger 4B models (Qwen3.5-4B, AgentDecision-4B, and both \data heads) plateau at 38--44\% due to missing room priors.

\begin{table}[t]
  \centering
  \input{tables/tab_offline}
  \caption{Decision-level accuracy (\%) on offline development sets. \emph{Dev} uses dual-teacher consensus labels on held-out container trajectories, whereas SOP and ALFWorld measure agreement with \stwo on development E decisions. Zero-shot Qwen rows average across four cyclic option rotations, pointer rows use one forward pass, and \data rows report three-seed means. Bold: column best.}
  \label{tab:offline}
\end{table}

\paragraph{Calibration objective analysis.}
On \emph{Dev}, our 4B \sone achieves lower negative log-likelihood (NLL) than commercial Jev and AgentDecision-4B (0.52 vs.\ 0.58 and 0.69) at 0.07 expected calibration error (ECE).
Replacing label smoothing ($\epsilon=0.1$) with hard-label cross-entropy ($\epsilon=0$) or CE + Brier leaves offline accuracy nearly unchanged (within 0.5 points; Table~\ref{tab:ablation}c) while pushing unsmoothed predictions toward probability $1.0$ even on unfamiliar states.
Consequently, the threshold fitted for 95\% agreement inflates near $1.0$ and reduces three-set mean gated agreement from 97.7\% to 92.4--93.7\%.

\paragraph{Confidence ranking vs.\ random delegation.}
Comparing our single-pass confidence gate ($c(x)\ge\tau$) against matched-rate random delegation across All-S2 E decisions (Table~\ref{tab:random}) confirms that uncertainty ranking—not lower delegation volume—drives accuracy, cutting AURC by $2\text{--}6\times$ and reducing ALFWorld disagreements from 16.0 to 3.8 per 100 decisions.

\begin{table}[t]
  \centering\footnotesize
  \setlength{\tabcolsep}{3pt}
  \resizebox{\columnwidth}{!}{%
  \begin{tabular}{@{}lccccc@{}}
  \toprule
  & \multicolumn{2}{c}{AURC $\downarrow$} & & \multicolumn{2}{c}{\gmain: agree / errors per 100} \\
  \cmidrule(lr){2-3}\cmidrule(lr){5-6}
  Benchmark & Gate & Random & Delegated & Gate & Random \\
  \midrule
  SOPBench & \textbf{0.014} & 0.086 & 98.0\% & \textbf{92.7 / 7.1} & 91.4 / 8.4 \\
  ALFWorld & \textbf{0.100} & 0.318 & 50.4\% & \textbf{92.4 / 3.8} & 68.2 / 16.0 \\
  WebShop & \textbf{0.156} & 0.302 & 41.4\% & \textbf{84.3 / 6.5} & 69.8 / 12.5 \\
  \bottomrule
  \end{tabular}%
  }
  \caption{Confidence gating versus matched-rate random delegation on All-S2 test E decisions. Reports risk--coverage AURC ($\downarrow$), online \gmain delegation share, and delegated \stwo agreement (\%) / disagreements per 100 E decisions. Bold: better per metric.}
  \label{tab:random}
\end{table}

\paragraph{Logged online dual-system execution trace.}
Figure~\ref{fig:online-trace} walks through a complete logged dual-system execution trace on WebShop \texttt{apparel}, illustrating how deterministic code (\textbf{D}), confident single-pass \sone delegation (\textbf{E}), gate-triggered \stwo deferral (\textbf{E} $\to$ \stwo), and open generation (\textbf{G}) interleave at runtime.

\begin{figure*}[t]
  \centering
  \begin{examplebox}{Online Dual-System Inference Trace (\texttt{WebShop/apparel} Episode 533, \gmain $\tau=0.948$, \stwo = DeepSeek-V4.1-Flash)}
    \erole{Task Goal:} \emph{``i want a navy fleece lined womens winter coat, and price lower than 90.00 dollars''}
    \tcbsep
    \begin{itemize}\setlength{\itemsep}{1.5pt}\setlength{\parskip}{0pt}\setlength{\parsep}{0pt}
      \item \textbf{[1] \texttt{G (s0)} $\rightarrow$ \stwo:} Generates concise query \texttt{"women's navy fleece lined winter coat"} (1 short \stwo call). $\rightarrow$ \textbf{\texttt{D (q1, inc1, r0)} [Code, 0 tok]:} Executes \texttt{search[...]} (50 results), increments \texttt{searches:=1}, verifies \texttt{unvisited\_left=True}.
      \item \textbf{[2] \texttt{E (ws/533/4, results screen)} $\rightarrow$ \sone Accepted ($p=0.955 \ge 0.948$):} Binary check whether results include the requested product kind; \sone outputs \texttt{yes} in one pass (0 \stwo tokens).
      \item \textbf{[3] \texttt{E (ws/533/7, pick)} $\rightarrow$ Gate Defers to \stwo ($p=0.358 < 0.948$):} Across 9 candidate ASINs + \texttt{none}, \sone's top score is uncertain (\texttt{b09mfb7nhz}: $0.358$, \texttt{none}: $0.198$). Gate escalates to \stwo, which avoids the $6.09$--$10.89$ cardigan and selects \texttt{b09jg8ywy3} (\$24.99). $\rightarrow$ \textbf{\texttt{D (open, item2)} [Code]:} Clicks \texttt{b09jg8ywy3}, checks $\$24.99 \le \$90.00$.
      \item \textbf{[4] \texttt{E (ws/533/11, match)} $\rightarrow$ \sone Accepted ($p=0.958 \ge 0.948$):} Confirms \emph{Goldweather Womens Winter Fleece Lined Coats... (Green, 3XL)} matches garment type and attributes $\to$ \sone outputs \texttt{yes}.
      \item \textbf{[5] \texttt{E (ws/533/13--17, variants)} $\rightarrow$ Mixed \sone/\stwo + \texttt{D (back)}:} On \texttt{size}, \sone picks \texttt{none} ($0.888 < 0.948 \to$ \stwo \texttt{none}); on \texttt{vgnone} (missing size?), \sone outputs \textbf{\texttt{no} ($p=0.953 \ge 0.948 \to$ \sone Accepted)}. On \texttt{color} (\{\texttt{beige, black, green, orange, none}\}; \texttt{navy} missing), \sone picks \texttt{none} ($0.894 \to$ \stwo) and \texttt{vgnone} outputs \texttt{yes} ($p=0.917 < 0.948 \to$ \stwo confirms \texttt{yes}). \textbf{\texttt{D (back)} [Code]:} Executes \texttt{click[< prev]}.
      \item \textbf{[6] \texttt{E (ws/533/20, pick)} $\rightarrow$ Gate Defers to \stwo ($p=0.274 < 0.948$):} \sone is uncertain over remaining candidates (\texttt{none}: $0.274$, \texttt{b09ky6ds6z}: $0.175$); \stwo selects \texttt{b09ky6ds6z} (\$36.99). $\rightarrow$ \textbf{\texttt{D (open, item2)} [Code]:} Opens product, checks $\$36.99 \le \$90.00$.
      \item \textbf{[7] \texttt{E (ws/533/24--29)} + \texttt{D (sel, buy)}:} \sone handles \texttt{match} (\textbf{\texttt{yes}, $p=0.955 \ge 0.948$}) and size \texttt{vgnone} (\textbf{\texttt{no}, $p=0.954 \ge 0.948$}); \stwo confirms unconstrained \texttt{size=none} ($0.877$) and selects \texttt{color=navy} ($p=0.892 < 0.948 \to$ \stwo agrees on \texttt{navy}). \textbf{\texttt{D (sel, pricecheck, buy)} [Code]:} Clicks \texttt{navy}, re-verifies $\$36.99 \le \$90.00$, executes \texttt{click[buy now]}.
    \end{itemize}
  \end{examplebox}
  \caption{Logged online dual-system trace on WebShop \texttt{apparel} Episode~533 under \gmain ($\tau=0.948$). Routine high-confidence checks run in one \sone pass, ambiguous product or attribute choices defer to \stwo, and navigation and budget checks execute as zero-token code (\textbf{D}).}
  \label{fig:online-trace}
\end{figure*}

%% file: tables/tab_offline.tex
\footnotesize
\setlength{\tabcolsep}{3.8pt}
\resizebox{\columnwidth}{!}{%
\begin{tabular}{@{}lcccc@{}}
\toprule
& \textbf{Dev} & \textbf{SOP} & \multicolumn{2}{c}{\textbf{ALFWorld}} \\
\cmidrule(lr){4-5}
Model & & E & E & search \\
\midrule
\multicolumn{5}{@{}l}{\textit{Zero-shot baselines}} \\
Kev-4B & 57.8 & 88.7 & 56.9 & 16.3 \\
AgentDecision-4B & 74.6 & 88.2 & 69.5 & 40.4 \\
Qwen3.5-4B & 71.4 & 91.7 & 70.9 & 44.0 \\
Qwen3.5-27B & 85.9 & 93.8 & 77.2 & 56.8 \\
Jev (commercial) & 78.4 & \textbf{95.5} & \textbf{81.2} & \textbf{66.4} \\
\midrule
\multicolumn{5}{@{}l}{\textit{Fine-tuned 4B baselines (4B-Base)}} \\
LLM-written synthetic (\syn) & 65.0 & 91.7 & 63.6 & 28.6 \\
\midrule
\multicolumn{5}{@{}l}{\textit{Fine-tuned on \data (4B-Base, 3 seeds)}} \\
Standard pointer head & 82.7 & 93.7 & 68.0 & 37.8 \\
Option-interaction head (\sone) & \textbf{83.8} & 93.5 & 68.9 & 41.5 \\
\bottomrule
\end{tabular}%
}

%% file: main.bbl
\begin{thebibliography}{64}
\providecommand{\natexlab}[1]{#1}

\bibitem[{{Anthropic}(2025)}]{anthropic2025skills}
{Anthropic}. 2025.
\newblock Equipping agents for the real world with agent skills.
\newblock
  \url{https://www.anthropic.com/engineering/equipping-agents-for-the-real-world-with-agent-skills}.

\bibitem[{Barres et~al.(2025)Barres, Dong, Ray, Si, and
  Narasimhan}]{barres2025tau2}
Victor Barres, Honghua Dong, Soham Ray, Xujie Si, and Karthik Narasimhan. 2025.
\newblock {$\tau^2$-Bench: Evaluating Conversational Agents in a Dual-Control
  Environment}.
\newblock \emph{arXiv preprint arXiv:2506.07982}.

\bibitem[{Cao et~al.(2026)Cao, Zuo, Papakostas, Xu, Xu, Meng, Zhou, Ren, Lyu,
  and Yan}]{cao2026searchjev}
Congfeng Cao, Lipeng Zuo, Konstantinos Papakostas, Qiwei Xu, Songwei Xu, Qianru
  Meng, Lun Zhou, Zhaochun Ren, Yougang Lyu, and Xiaohui Yan. 2026.
\newblock {SearchJev: A Calibrated System-1 Model for Search Agents}.
\newblock \emph{arXiv preprint arXiv:2610.05107}.

\bibitem[{Chen et~al.(2026)Chen, Feng, Xia, and Chen}]{chen2026skvm}
Le~Chen, Erhu Feng, Yubin Xia, and Haibo Chen. 2026.
\newblock {SkVM: Revisiting Language VM for Skills across Heterogenous LLMs and
  Harnesses}.
\newblock \emph{arXiv preprint arXiv:2604.03088}.

\bibitem[{Chen et~al.(2024)Chen, Zaharia, and Zou}]{chen2023frugalgpt}
Lingjiao Chen, Matei Zaharia, and James Zou. 2024.
\newblock {FrugalGPT: How to Use Large Language Models While Reducing Cost and
  Improving Performance}.
\newblock \emph{Transactions on Machine Learning Research}.
\newblock ArXiv:2305.05176.

\bibitem[{Cho and Park(2026)}]{cho2026skillfollowing}
Seonghyeon Cho and Chanjun Park. 2026.
\newblock {Skill Following: Evaluating Actual Skill Use in Retrieval-Enabled
  LLM Agents}.
\newblock In \emph{Findings of the Association for Computational Linguistics:
  EMNLP 2026}.
\newblock ArXiv:2609.00549.

\bibitem[{{DeepSeek-AI}(2026)}]{deepseek2026v4}
{DeepSeek-AI}. 2026.
\newblock {DeepSeek-V4: Towards Highly Efficient Million-Token Context
  Intelligence}.
\newblock \emph{arXiv preprint arXiv:2606.19348}.

\bibitem[{Dong et~al.(2026)Dong, Gao, Li, Xu, Hua, and Yang}]{dong2026harmful}
Gen Dong, Yanjie Gao, Liqun Li, Tianyin Xu, Yu~Hua, and Fan Yang. 2026.
\newblock {Agent Skills Can Be Harmful: An Empirical Study of Skill-Induced
  Failures in LLM Agents}.
\newblock \emph{arXiv preprint arXiv:2608.11888}.

\bibitem[{{FayeZC}(2026)}]{skillmd138k}
{FayeZC}. 2026.
\newblock {SkillMD-138K}: A dataset of agent skill files.
\newblock Hugging Face dataset,
  \url{https://huggingface.co/datasets/FayeZC/SkillMD-138K}.

\bibitem[{Geifman and El-Yaniv(2017)}]{geifman2017selective}
Yonatan Geifman and Ran El-Yaniv. 2017.
\newblock {Selective Classification for Deep Neural Networks}.
\newblock In \emph{Advances in Neural Information Processing Systems}.
\newblock ArXiv:1705.08500.

\bibitem[{Guo et~al.(2017)Guo, Pleiss, Sun, and
  Weinberger}]{guo2017calibration}
Chuan Guo, Geoff Pleiss, Yu~Sun, and Kilian~Q. Weinberger. 2017.
\newblock {On Calibration of Modern Neural Networks}.
\newblock In \emph{Proceedings of the 34th International Conference on Machine
  Learning}.
\newblock ArXiv:1706.04599.

\bibitem[{Hemadri et~al.(2026)Hemadri, Mohammed, Maheshwary, Daruru, Davasam,
  Yadav, Sunkara, and Rajeswar}]{hemadri2026r2v}
Raghu~Vamshi Hemadri, Humaira~Firdowse Mohammed, Rishabh Maheshwary, Srivatsava
  Daruru, Sagar Davasam, Vikas Yadav, Srinivas Sunkara, and Sai Rajeswar. 2026.
\newblock {R2V Agent: Teaching SLMs When to Ask for Help}.
\newblock \emph{arXiv preprint arXiv:2605.16604}.

\bibitem[{Hu et~al.(2022)Hu, Shen, Wallis, Allen-Zhu, Li, Wang, Wang, and
  Chen}]{hu2022lora}
Edward~J. Hu, Yelong Shen, Phillip Wallis, Zeyuan Allen-Zhu, Yuanzhi Li, Shean
  Wang, Lu~Wang, and Weizhu Chen. 2022.
\newblock {LoRA: Low-Rank Adaptation of Large Language Models}.
\newblock In \emph{International Conference on Learning Representations}.
\newblock ArXiv:2106.09685.

\bibitem[{Kahneman(2011)}]{kahneman2011thinking}
Daniel Kahneman. 2011.
\newblock \emph{Thinking, Fast and Slow}.
\newblock Farrar, Straus and Giroux, New York.

\bibitem[{Kapoor et~al.(2025)Kapoor, Stroebl, Siegel, Nadgir, and
  Narayanan}]{kapoor2024agentsmatter}
Sayash Kapoor, Benedikt Stroebl, Zachary~S. Siegel, Nitya Nadgir, and Arvind
  Narayanan. 2025.
\newblock {AI Agents That Matter}.
\newblock \emph{Transactions on Machine Learning Research}.
\newblock ArXiv:2407.01502.

\bibitem[{Khattab et~al.(2024)Khattab, Singhvi, Maheshwari, Zhang, Santhanam,
  Vardhamanan, Haq, Sharma, Joshi, Moazam, Miller, Zaharia, and
  Potts}]{khattab2024dspy}
Omar Khattab, Arnav Singhvi, Paridhi Maheshwari, Zhiyuan Zhang, Keshav
  Santhanam, Sri Vardhamanan, Saiful Haq, Ashutosh Sharma, Thomas~T. Joshi,
  Hanna Moazam, Heather Miller, Matei Zaharia, and Christopher Potts. 2024.
\newblock {DSPy: Compiling Declarative Language Model Calls into Self-Improving
  Pipelines}.
\newblock In \emph{The Twelfth International Conference on Learning
  Representations}.
\newblock ArXiv:2310.03714.

\bibitem[{Kwa et~al.(2025)Kwa, West, Becker, Deng, Garcia, Hasin, Jawhar,
  Kinniment, Rush, Arx, Bloom, Broadley, Du, Goodrich, Jurkovic, Miles, Nix,
  Lin, Painter, Parikh, Rein, Sato, Wijk, Ziegler, Barnes, and
  Chan}]{kwa2025longtasks}
Thomas Kwa, Ben West, Joel Becker, Amy Deng, Katharyn Garcia, Max Hasin, Sami
  Jawhar, Megan Kinniment, Nate Rush, Sydney~Von Arx, Ryan Bloom, Thomas
  Broadley, Haoxing Du, Brian Goodrich, Nikola Jurkovic, Luke~Harold Miles,
  Seraphina Nix, Tao Lin, Chris Painter, and 7 others. 2025.
\newblock {Measuring AI Ability to Complete Long Software Tasks}.
\newblock In \emph{Advances in Neural Information Processing Systems}.
\newblock ArXiv:2503.14499.

\bibitem[{Li(2026)}]{li2026hexis}
Minghao Li. 2026.
\newblock {HEXIS: Compiling Agent Skills into Extended Finite State Machines}.
\newblock \emph{arXiv preprint arXiv:2609.30123}.

\bibitem[{Li et~al.(2026{\natexlab{a}})Li, Liu, Chen, You, Di, He, Zheng, Choe,
  Sun, Wang, Tao, Li, Zhao, Geng, Wu, Zhou, Chen, Xing, Li, Zeng, Wang, Wang,
  Chaim, Jiang, Shen, Kong, Liu, Wang, Liu, Li, Lan, Lin, Ye, He, Li, Zhang,
  Gao, Li, Ma, Jing, Wang, Li, Xue, Lyu, He, Tian, Wu, Wang, Gao, Chen, Liu,
  Cheng, Bao, Tong, Xu, Zhuo, Ye, Qi, Li, Liao, Tan, Shi, Tang, Tankasala,
  Yuan, Qian, Tu, Wang, Sun, Wang, Taylor, Yang, Guan, Dong, Zhang, Dillmann,
  chung Lee, and Song}]{li2026skillsbench}
Xiangyi Li, Yimin Liu, Wenbo Chen, Bingran You, Zonglin Di, Yifeng He, Shenghan
  Zheng, Kyoung~Whan Choe, Jiankai Sun, Shuyi Wang, Chujun Tao, Binxu Li,
  Xuandong Zhao, Hejia Geng, Xiaojun Wu, Junwei Zhou, Xiaokun Chen, Hanwen
  Xing, Yubo Li, and 59 others. 2026{\natexlab{a}}.
\newblock {SkillsBench: Benchmarking How Well Agent Skills Work Across Diverse
  Tasks}.
\newblock \emph{arXiv preprint arXiv:2602.12670}.

\bibitem[{Li et~al.(2026{\natexlab{b}})Li, Krishnan, and
  Padman}]{li2026readingright}
Yubo Li, Ramayya Krishnan, and Rema Padman. 2026{\natexlab{b}}.
\newblock {Reading Right, Acting Wrong: Diagnosing Decision Failures of System
  One Models in Agent Loops}.
\newblock \emph{arXiv preprint arXiv:2610.18579}.

\bibitem[{Li et~al.(2026{\natexlab{c}})Li, Huang, Wang, Zhang, Antoniades, Hua,
  Zhu, Zeng, Wang, Wang, and Yan}]{li2025sopbench}
Zekun Li, Shinda Huang, Jiangtian Wang, Nathan Zhang, Antonis Antoniades,
  Wenyue Hua, Kaijie Zhu, Sirui Zeng, Chi Wang, William~Yang Wang, and Xifeng
  Yan. 2026{\natexlab{c}}.
\newblock {SOPBench: Evaluating Language Agents at Following Standard Operating
  Procedures and Constraints}.
\newblock In \emph{Proceedings of the 2026 Conference on Empirical Methods in
  Natural Language Processing}.
\newblock ArXiv:2503.08669.

\bibitem[{Lin et~al.(2023)Lin, Fu, Yang, Brahman, Huang, Bhagavatula,
  Ammanabrolu, Choi, and Ren}]{lin2023swiftsage}
Bill~Yuchen Lin, Yicheng Fu, Karina Yang, Faeze Brahman, Shiyu Huang, Chandra
  Bhagavatula, Prithviraj Ammanabrolu, Yejin Choi, and Xiang Ren. 2023.
\newblock {SwiftSage: A Generative Agent with Fast and Slow Thinking for
  Complex Interactive Tasks}.
\newblock In \emph{Advances in Neural Information Processing Systems}.
\newblock ArXiv:2305.17390.

\bibitem[{Liu et~al.(2026)Liu, Xu, Wang, Gao, and Wang}]{liu2026harecap}
Shen Liu, Zhenguo Xu, Shaopu Wang, Yike Gao, and Chunlei Wang. 2026.
\newblock {HaReCAP: Habitual-action Grounding for Recursive Large Language
  Model Agents}.
\newblock \emph{arXiv preprint arXiv:2608.16447}.

\bibitem[{Lu et~al.(2026)Lu, Yao, Wu, Han, Gu, Cai, Lu, Xiao, Zhuang, and
  Shen}]{lu2026skill0}
Zhengxi Lu, Zhiyuan Yao, Jinyang Wu, Chengcheng Han, Qi~Gu, Xunliang Cai,
  Weiming Lu, Jun Xiao, Yueting Zhuang, and Yongliang Shen. 2026.
\newblock {SKILL0: In-Context Agentic Reinforcement Learning for Skill
  Internalization}.
\newblock In \emph{Findings of the Association for Computational Linguistics:
  EMNLP 2026}.
\newblock ArXiv:2604.02268.

\bibitem[{Lukassen et~al.(2026)Lukassen, Weisser, and
  Silbersdorff}]{lukassen2026cheap}
Fabian Lukassen, Christoph Weisser, and Alexander Silbersdorff. 2026.
\newblock {Cheap, Consistent, Calibrated, but Biased? Typed Decision Models as
  Judges}.
\newblock \emph{arXiv preprint arXiv:2610.18122}.

\bibitem[{Mazza and Levi(2026)}]{mazza2026scope}
Arnon Mazza and Elad Levi. 2026.
\newblock {SCOPE: Compiling Natural-Language Policies into Neurosymbolic
  Predicate Programs}.
\newblock \emph{arXiv preprint arXiv:2610.31062}.

\bibitem[{Müller et~al.(2019)Müller, Kornblith, and
  Hinton}]{muller2019labelsmoothing}
Rafael Müller, Simon Kornblith, and Geoffrey Hinton. 2019.
\newblock {When Does Label Smoothing Help?}
\newblock In \emph{Advances in Neural Information Processing Systems}.
\newblock ArXiv:1906.02629.

\bibitem[{Ong et~al.(2025)Ong, Almahairi, Wu, Chiang, Wu, Gonzalez, Kadous, and
  Stoica}]{ong2025routellm}
Isaac Ong, Amjad Almahairi, Vincent Wu, Wei-Lin Chiang, Tianhao Wu, Joseph~E.
  Gonzalez, M~Waleed Kadous, and Ion Stoica. 2025.
\newblock {RouteLLM: Learning to Route LLMs with Preference Data}.
\newblock In \emph{The Thirteenth International Conference on Learning
  Representations}.
\newblock ArXiv:2406.18665.

\bibitem[{Palmer(2026)}]{kev2026}
Jared Palmer. 2026.
\newblock {Kev}: Open typed decision models.
\newblock \url{https://github.com/jaredpalmer/kev}.

\bibitem[{Prasad et~al.(2024)Prasad, Koller, Hartmann, Clark, Sabharwal,
  Bansal, and Khot}]{prasad2024adapt}
Archiki Prasad, Alexander Koller, Mareike Hartmann, Peter Clark, Ashish
  Sabharwal, Mohit Bansal, and Tushar Khot. 2024.
\newblock {ADaPT: As-Needed Decomposition and Planning with Language Models}.
\newblock In \emph{Findings of the Association for Computational Linguistics:
  NAACL 2024}.

\bibitem[{{Qwen Team}(2026)}]{qwen2026qwen35}
{Qwen Team}. 2026.
\newblock {Qwen3.5}.
\newblock Model card, \url{https://huggingface.co/Qwen/Qwen3.5-4B-Base}.

\bibitem[{Saha et~al.(2025)Saha, Prasad, Chen, Hase, Stengel-Eskin, and
  Bansal}]{saha2024system1x}
Swarnadeep Saha, Archiki Prasad, Justin Chih-Yao Chen, Peter Hase, Elias
  Stengel-Eskin, and Mohit Bansal. 2025.
\newblock {System-1.x: Learning to Balance Fast and Slow Planning with Language
  Models}.
\newblock In \emph{The Thirteenth International Conference on Learning
  Representations}.
\newblock ArXiv:2407.14414.

\bibitem[{Shridhar et~al.(2021)Shridhar, Yuan, Côté, Bisk, Trischler, and
  Hausknecht}]{shridhar2021alfworld}
Mohit Shridhar, Xingdi Yuan, Marc-Alexandre Côté, Yonatan Bisk, Adam
  Trischler, and Matthew Hausknecht. 2021.
\newblock {ALFWorld: Aligning Text and Embodied Environments for Interactive
  Learning}.
\newblock In \emph{International Conference on Learning Representations}.
\newblock ArXiv:2010.03768.

\bibitem[{Song et~al.(2024)Song, Yin, Yue, Huang, Li, and Lin}]{song2024eto}
Yifan Song, Da~Yin, Xiang Yue, Jie Huang, Sujian Li, and Bill~Yuchen Lin. 2024.
\newblock {Trial and Error: Exploration-Based Trajectory Optimization of LLM
  Agents}.
\newblock In \emph{Proceedings of the 62nd Annual Meeting of the Association
  for Computational Linguistics (Volume 1: Long Papers)}.

\bibitem[{Sowietzki et~al.(2026)Sowietzki, Holter, Metz, Kasper, Baur, and
  El-Assady}]{sowietzki2026aria}
Pascal Sowietzki, Steffen Holter, Yannick Metz, Emil Kasper, Raphael Baur, and
  Mennatallah El-Assady. 2026.
\newblock {Ask Your Advisor: Learning When to Consult Stronger Models in
  Agentic Systems}.
\newblock \emph{arXiv preprint arXiv:2610.13870}.

\bibitem[{Su et~al.(2025)Su, Sukhbaatar, Rabbat, Tian, and
  Zheng}]{su2024dualformer}
DiJia Su, Sainbayar Sukhbaatar, Michael Rabbat, Yuandong Tian, and Qinqing
  Zheng. 2025.
\newblock {Dualformer: Controllable Fast and Slow Thinking by Learning with
  Randomized Reasoning Traces}.
\newblock In \emph{The Thirteenth International Conference on Learning
  Representations}.
\newblock ArXiv:2410.09918.

\bibitem[{Suh et~al.(2026)Suh, Nguyen, Lee, Hartman, Liu, Wang, Li, and
  Yang}]{suh2026nscu}
Hyewon Suh, Thanh~Minh Nguyen, Chih-Lun Lee, Darrow Hartman, Lizhao Liu,
  Xin~Eric Wang, Ang Li, and Jiachen Yang. 2026.
\newblock {Neuro-Symbolic Computer Use: Learning Reusable Policies for Reliable
  and Efficient Execution}.
\newblock \emph{arXiv preprint arXiv:2609.36927}.

\bibitem[{Sun et~al.(2026)Sun, Xu, Shi, and Yang}]{sun2026typesafe}
Yu~Sun, Junhao Xu, Jiajia Shi, and Zijin Yang. 2026.
\newblock {Type-Safe Is Not Error-Free: Typed Decision Models Follow the Option
  Name, Not the Definition Bound to It}.
\newblock \emph{arXiv preprint arXiv:2609.26758}.

\bibitem[{Tan et~al.(2026)Tan, Zhang, Li, Cui, Geng, Zhang, Zhang, Chen, Wang,
  Wang, Yin, Hu, Zhang, and Bai}]{tan2026skt}
Zelin Tan, Yiqun Zhang, Hao Li, Zhiyao Cui, Hejia Geng, Shao Zhang, Hangfan
  Zhang, Yang Chen, Xiaosong Wang, Lilong Wang, Zhenfei Yin, Shuyue Hu, Chen
  Zhang, and Lei Bai. 2026.
\newblock {SKT: Skill-Use Training at Scale via Verified Synthetic Data
  Generation}.
\newblock \emph{arXiv preprint arXiv:2608.02287}.

\bibitem[{{TypeSafe AI}(2026)}]{typesafe2026jev}
{TypeSafe AI}. 2026.
\newblock {Jev}: A system one model.
\newblock Model documentation, \url{https://docs.typesafe.ai}. Version
  \texttt{jev-1.14.0}.

\bibitem[{Vats(2026)}]{vats2026goodjudgment}
Naman Vats. 2026.
\newblock {Good Judgment, Bad Advice: Decision-Model Controllers Raise Agent
  Cost Without Improving Success}.
\newblock \emph{arXiv preprint arXiv:2610.24649}.

\bibitem[{Wang et~al.(2024)Wang, Xie, Jiang, Mandlekar, Xiao, Zhu, Fan, and
  Anandkumar}]{wang2023voyager}
Guanzhi Wang, Yuqi Xie, Yunfan Jiang, Ajay Mandlekar, Chaowei Xiao, Yuke Zhu,
  Linxi Fan, and Anima Anandkumar. 2024.
\newblock {Voyager: An Open-Ended Embodied Agent with Large Language Models}.
\newblock \emph{Transactions on Machine Learning Research}.
\newblock ArXiv:2305.16291.

\bibitem[{Wang et~al.(2026{\natexlab{a}})Wang, Hee, Koto, Baldwin, and
  Li}]{wang2026skillgym}
Renxi Wang, Mingshan Hee, Fajri Koto, Timothy Baldwin, and Haonan Li.
  2026{\natexlab{a}}.
\newblock {SkillGym: Training Skill-Use Agents with Automatic Verifiable
  Environment Generation}.
\newblock \emph{arXiv preprint arXiv:2609.37539}.

\bibitem[{Wang et~al.(2025)Wang, Li, Song, Xu, Tang, Zhuge, Pan, Song, Li,
  Singh, Tran, Li, Ma, Zheng, Qian, Shao, Muennighoff, Zhang, Hui, Lin,
  Brennan, Peng, Ji, and Neubig}]{wang2024openhands}
Xingyao Wang, Boxuan Li, Yufan Song, Frank~F. Xu, Xiangru Tang, Mingchen Zhuge,
  Jiayi Pan, Yueqi Song, Bowen Li, Jaskirat Singh, Hoang~H. Tran, Fuqiang Li,
  Ren Ma, Mingzhang Zheng, Bill Qian, Yanjun Shao, Niklas Muennighoff, Yizhe
  Zhang, Binyuan Hui, and 5 others. 2025.
\newblock {OpenHands: An Open Platform for AI Software Developers as Generalist
  Agents}.
\newblock In \emph{The Thirteenth International Conference on Learning
  Representations}.
\newblock ArXiv:2407.16741.

\bibitem[{Wang et~al.(2026{\natexlab{b}})Wang, Yao, Sun, Hu, Xiao, Luo, Han,
  Chen, Sun, and Deng}]{wang2026skillcorpus}
Yanze Wang, Pengfei Yao, Tianyi Sun, Chuanrui Hu, Yan Xiao, Xiaotian Luo,
  Yunyun Han, Yifan Chen, Jun Sun, and Yafeng Deng. 2026{\natexlab{b}}.
\newblock {SkillCorpus: Consolidating and Evaluating the Open Skill Ecosystem
  for Real-World LLM Agents}.
\newblock \emph{arXiv preprint arXiv:2607.15557}.

\bibitem[{Wang and Zang(2026)}]{wang2026abacus}
Yuhao Wang and Zelin Zang. 2026.
\newblock {ABACUS: Program-Guided Multistep Reasoning over Frozen Judgment
  Models}.
\newblock \emph{arXiv preprint arXiv:2610.25963}.

\bibitem[{Weber and Taneja(2026)}]{weber2026habits}
Travis Weber and Rohit Taneja. 2026.
\newblock {Making Agents More Consistent: Skills Should Form Habits for Repeat
  Tasks}.
\newblock \emph{arXiv preprint arXiv:2609.25299}.

\bibitem[{Wu and Lim(2026)}]{wu2026reflex}
Tiantong Wu and Wei Yang~Bryan Lim. 2026.
\newblock {REFLEX with Jev for Efficient Selective Control in LLM Agents}.
\newblock \emph{arXiv preprint arXiv:2609.26532}.

\bibitem[{Xia et~al.(2026)Xia, Chen, Wang, Liu, Zeng, Wang, Han, Zhou, Zhao,
  Chen, Zheng, Xie, and Yao}]{xia2026skillrl}
Peng Xia, Jianwen Chen, Hanyang Wang, Jiaqi Liu, Kaide Zeng, Yu~Wang, Siwei
  Han, Yiyang Zhou, Xujiang Zhao, Haifeng Chen, Zeyu Zheng, Cihang Xie, and
  Huaxiu Yao. 2026.
\newblock {SkillRL: Evolving Agents via Recursive Skill-Augmented Reinforcement
  Learning}.
\newblock In \emph{Advances in Neural Information Processing Systems}.
\newblock ArXiv:2602.08234.

\bibitem[{Xiang et~al.(2026)Xiang, Chu, Xu, Zhang, and Zhang}]{xiang2026las}
Dawei Xiang, Kexin Chu, Wenyan Xu, Wenhui Zhang, and Wei Zhang. 2026.
\newblock {LLM-as-Scheduler: Agentic Workflow Dynamic Scheduling}.
\newblock In \emph{Proceedings of the 64th Annual Meeting of the Association
  for Computational Linguistics (Volume 1: Long Papers)}.

\bibitem[{Xie et~al.(2024)Xie, Zhang, Chen, Li, Zhao, Cao, Hua, Cheng, Shin,
  Lei, Liu, Xu, Zhou, Savarese, Xiong, Zhong, and Yu}]{xie2024osworld}
Tianbao Xie, Danyang Zhang, Jixuan Chen, Xiaochuan Li, Siheng Zhao, Ruisheng
  Cao, Toh~Jing Hua, Zhoujun Cheng, Dongchan Shin, Fangyu Lei, Yitao Liu,
  Yiheng Xu, Shuyan Zhou, Silvio Savarese, Caiming Xiong, Victor Zhong, and Tao
  Yu. 2024.
\newblock {OSWorld: Benchmarking Multimodal Agents for Open-Ended Tasks in Real
  Computer Environments}.
\newblock In \emph{Advances in Neural Information Processing Systems}.
\newblock ArXiv:2404.07972.

\bibitem[{Xu et~al.(2026{\natexlab{a}})Xu, Chen, Pan, Guan, Dong, Li, and
  Pu}]{xu2026skillsmith}
Duling Xu, Zheng Chen, Zaifeng Pan, Jiawei Guan, Dong Dong, Jialin Li, and
  Bangzheng Pu. 2026{\natexlab{a}}.
\newblock {SkillSmith: Compiling Agent Skills into Boundary-Guided Runtime
  Interfaces}.
\newblock \emph{arXiv preprint arXiv:2605.15215}.

\bibitem[{Xu et~al.(2026{\natexlab{b}})Xu, Luo, and Feng}]{xu2026gengate}
Yanqiu Xu, Han Luo, and Lei Feng. 2026{\natexlab{b}}.
\newblock {Generation-Gated Cascaded Routing for Large Language Models}.
\newblock In \emph{Proceedings of the 2026 Conference on Empirical Methods in
  Natural Language Processing}.

\bibitem[{Yang et~al.(2024)Yang, Jimenez, Wettig, Lieret, Yao, Narasimhan, and
  Press}]{yang2024sweagent}
John Yang, Carlos~E. Jimenez, Alexander Wettig, Kilian Lieret, Shunyu Yao,
  Karthik Narasimhan, and Ofir Press. 2024.
\newblock {SWE-agent: Agent-Computer Interfaces Enable Automated Software
  Engineering}.
\newblock In \emph{Advances in Neural Information Processing Systems}.
\newblock ArXiv:2405.15793.

\bibitem[{Yao et~al.(2022)Yao, Chen, Yang, and Narasimhan}]{yao2022webshop}
Shunyu Yao, Howard Chen, John Yang, and Karthik Narasimhan. 2022.
\newblock {WebShop: Towards Scalable Real-World Web Interaction with Grounded
  Language Agents}.
\newblock In \emph{Advances in Neural Information Processing Systems}.
\newblock ArXiv:2207.01206.

\bibitem[{Yao et~al.(2025)Yao, Shinn, Razavi, and Narasimhan}]{yao2024taubench}
Shunyu Yao, Noah Shinn, Pedram Razavi, and Karthik Narasimhan. 2025.
\newblock {$\tau$-bench: A Benchmark for Tool-Agent-User Interaction in
  Real-World Domains}.
\newblock In \emph{The Thirteenth International Conference on Learning
  Representations}.
\newblock ArXiv:2406.12045.

\bibitem[{Yao et~al.(2023)Yao, Zhao, Yu, Du, Shafran, Narasimhan, and
  Cao}]{yao2023react}
Shunyu Yao, Jeffrey Zhao, Dian Yu, Nan Du, Izhak Shafran, Karthik Narasimhan,
  and Yuan Cao. 2023.
\newblock {ReAct: Synergizing Reasoning and Acting in Language Models}.
\newblock In \emph{The Eleventh International Conference on Learning
  Representations}.
\newblock ArXiv:2210.03629.

\bibitem[{Yin et~al.(2024)Yin, Brahman, Ravichander, Chandu, Chang, Choi, and
  Lin}]{yin2024lumos}
Da~Yin, Faeze Brahman, Abhilasha Ravichander, Khyathi Chandu, Kai-Wei Chang,
  Yejin Choi, and Bill~Yuchen Lin. 2024.
\newblock {Agent Lumos: Unified and Modular Training for Open-Source Language
  Agents}.
\newblock In \emph{Proceedings of the 62nd Annual Meeting of the Association
  for Computational Linguistics (Volume 1: Long Papers)}.

\bibitem[{Yu et~al.(2026)Yu, Yin, Fan, Yu, Zhong, and Li}]{yu2026compile}
Chenglin Yu, Li~Yin, Qingxin Fan, Ying Yu, RunyangRay Zhong, and Ming Li. 2026.
\newblock {Compile, Then Page: Executable SOP Programs and a Capability-Gated
  Runtime for Procedural LLM Agents}.
\newblock \emph{arXiv preprint arXiv:2607.11346}.

\bibitem[{Zeng et~al.(2024)Zeng, Liu, Lu, Wang, Liu, Dong, and
  Tang}]{zeng2024agenttuning}
Aohan Zeng, Mingdao Liu, Rui Lu, Bowen Wang, Xiao Liu, Yuxiao Dong, and Jie
  Tang. 2024.
\newblock {AgentTuning: Enabling Generalized Agent Abilities for LLMs}.
\newblock In \emph{Findings of the Association for Computational Linguistics:
  ACL 2024}.

\bibitem[{Zhang et~al.(2026)Zhang, Yang, Shi, Chen, Nie, Wan, Hong, Liu, and
  Wu}]{zhang2026anyjev}
Jiamu Zhang, Tianze Yang, Yucheng Shi, Evan Chen, Zixiang Nie, Kelly Wan,
  Liangjie Hong, Ninghao Liu, and Liang Wu. 2026.
\newblock {AnyJev Technical Report}.
\newblock \emph{arXiv preprint arXiv:2610.00831}.

\bibitem[{Zheng et~al.(2024)Zheng, Zhou, Meng, Zhou, and Huang}]{zheng2024mcq}
Chujie Zheng, Hao Zhou, Fandong Meng, Jie Zhou, and Minlie Huang. 2024.
\newblock {Large Language Models Are Not Robust Multiple Choice Selectors}.
\newblock In \emph{The Twelfth International Conference on Learning
  Representations}.
\newblock ArXiv:2309.03882.

\bibitem[{Zhou et~al.(2024)Zhou, Xu, Zhu, Zhou, Lo, Sridhar, Cheng, Ou, Bisk,
  Fried, Alon, and Neubig}]{zhou2024webarena}
Shuyan Zhou, Frank~F. Xu, Hao Zhu, Xuhui Zhou, Robert Lo, Abishek Sridhar,
  Xianyi Cheng, Tianyue Ou, Yonatan Bisk, Daniel Fried, Uri Alon, and Graham
  Neubig. 2024.
\newblock {WebArena: A Realistic Web Environment for Building Autonomous
  Agents}.
\newblock In \emph{International Conference on Learning Representations}.
\newblock ArXiv:2307.13854.

\bibitem[{Zhu et~al.(2026)Zhu, Feng, Han, Wang, Yang, Tang, You, and
  Ji}]{zhu2026jevagentharness}
Kunlun Zhu, Tao Feng, Xu~Han, Heng Wang, Ling Yang, Xiangru Tang, Jiaxuan You,
  and Heng Ji. 2026.
\newblock {JevAgentHarness: Benchmarking, Deploying and Training Decision
  Models That Share Control with LLMs}.
\newblock \emph{arXiv preprint arXiv:2610.21412}.

\end{thebibliography}
